\documentclass{bmvc2k}

\usepackage{times}
\usepackage{epsfig}
\usepackage{graphicx}
\usepackage{amsmath}
\usepackage{amssymb}
\usepackage{subcaption}
\usepackage{multirow}
\usepackage{adjustbox}
\usepackage[font=footnotesize]{caption}
\usepackage[dvipsnames]{xcolor}
\definecolor{sec_title_color}{rgb}{0,.1,.4}

\title{Efficient Coarse-to-Fine Non-Local Module\\ for the
	Detection of Small Objects}

\addauthor{Hila Levi}{hila.levi@weizmann.ac.il}{1}
\addauthor{Shimon Ullman}{shimon.ullman@weizmann.ac.il}{1}

\addinstitution{
Dept. of Computer Science and Applied Mathematics,
The Weizmann Institute of Science, Israel
}

\runninghead{}{}

\begin{document}

\maketitle
\vspace*{-5mm}
\begin{figure}[h]
	\begin{center}
		\begin{subfigure}[b]{0.49\linewidth}
			\adjincludegraphics[width=\linewidth,trim={{.35\width} {.05\height} 0 {.32\height}},clip,keepaspectratio]{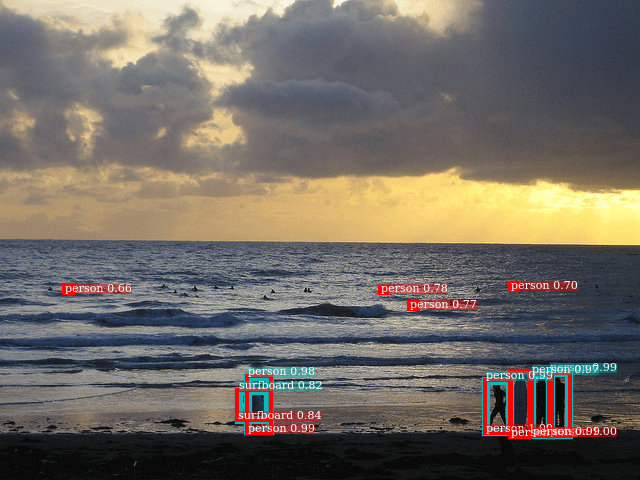}
			\caption{Can you see the swimming people?}
			\label{fig:intro2}
		\end{subfigure}
		\begin{subfigure}[b]{0.49\linewidth}
			\adjincludegraphics[width=\linewidth,trim={{.15\width} {.15\height} {.13\width} {.103\height}},clip,keepaspectratio]{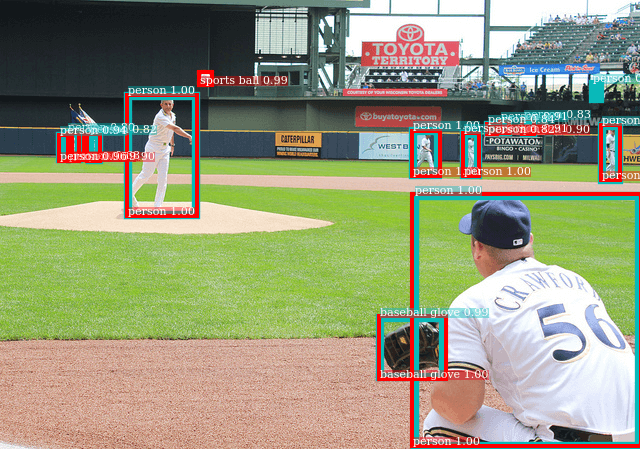}
			\caption{Where is the ball?}
			\label{fig:intro1}
		\end{subfigure}
	\end{center}
	\caption{Using a relational non-local module directly on the feature maps in a coarse-to-fine manner enables the detection of small objects, based on (i) repeating instances of the same class and (ii) the existence of larger related objects, allowing us to: (a) pay attention to the tiny swimmers in the sea and (b) locate the ball. Cyan - NL, Red - ours, ENL. Best viewed in color.}
	\label{fig:introduction}
\end{figure}

\begin{abstract}
An image is not just a collection of objects, but rather a graph where each object is related to other objects through spatial and semantic relations. Using relational reasoning modules, such as the non-local module \cite{wang2017non},  can therefore improve object detection. Current schemes apply such dedicated modules either to a specific layer of the bottom-up stream, or between already-detected objects. We show that the relational process can be better modeled in a coarse-to-fine manner and  present a novel framework, applying a non-local module sequentially to increasing resolution feature maps along the top-down stream. In this way, information can naturally passed from larger objects to smaller related ones. Applying the module to fine feature maps further allows the information to pass between the small objects themselves, exploiting repetitions of instances of the same class. In practice, due to the expensive memory utilization of the non-local module, it is infeasible to apply the module as currently used to high-resolution feature maps. We redesigned the non local module, improved it in terms of memory and number of operations, allowing it to be placed anywhere along the network. We further incorporated  relative spatial information into the module, in a manner that can be incorporated into our efficient implementation. We show the effectiveness of our scheme by improving the results of detecting small objects on COCO by 1-2 AP points over Faster and Mask RCNN and by 1 AP over using non-local module on the bottom-up stream. 
\end{abstract}

\section{Introduction}
\label{sec:intro}
Scene understanding has shown an impressing improvement in the last few years. Since the revival of deep neural networks, there has been a significant increase in the performance of a range of relevant tasks, including classification, object detection, segmentation, part localization etc.

Early works relied heavily on the hierarchical structure of bottom-up classification networks to perform additional tasks such as detection \cite{girshick2014rich, girshick2015fast, redmon2016you, he2014spatial}, by using the last network layer to predict object locations. A next significant step, partly motivated by the human vision system, incorporated context into the detection scheme by using a bottom-up top-down architecture \cite{lin2017feature, he2017mask, fu2017dssd, redmon2018yolov3}. This architecture combines high level contextual data from the last layers with highly localized fine-grained information expressed in lower layers. A further challenge, which became an active research area, is to incorporate relational reasoning into the detection systems \cite{battaglia2018relational, chen2018iterative, raposo2017discovering}. By using relational reasoning, an image forms not just a collection of unrelated objects, but rather resembling a "scene graph" \cite{johnson2015image} of entities (nodes, objects) connected by edges (relations, predicates). 

In this line of development, the detection of small objects still remains a difficult task. This task was shown to benefit from the use of context \cite{divvala2009empirical, galleguillos2010context}, and the current work applies the use of relations as context for the detection of small objects. 
Consider for example the images in figure \ref{fig:introduction}. The repetition of instances from the same class in the image, as well as the existence of larger instances from related classes, serve as semantic detection clues. It enables the detection of the tiny people in the sea (figure \ref{fig:introduction}a), partly based on the existence of the larger people in the shore. It similarly localizes the small sport ball, partly based on the throwing man and the player's glove (figure \ref{fig:introduction}b).  

Exploiting relations information, specifically for small object detection, requires propagating information over large distances in high resolution feature maps according to the data in a specific image. This is difficult to achieve by convolutional layers, 
since they transmit information over short distances only and in the same manner for all images, based on learning.
Recently, a Non-Local module \cite{wang2017non} has been formulated and integrated into CNNs \cite{vaswani2017attention, santoro2017simple}. 
The non-local module is capable to pass information between distant pixels according to their appearance, and is applicable to our current task. 
Using it sequentially, in a coarse-to-fine manner, enables to pass semantic information from larger, easy to detect objects, to smaller ones. Using the non-local (NL) module in lower layers allows information propagation between the small objects themselves.

For the current needs, there are two disadvantages in the original design of the NL module. The first is its expensive computational and memory budget. 
In preceding works this block was integrated into high layers of the bottom-up stream, but in our task it is integrated into lower-level layers, where its memory demands become infeasible. Furthermore, in detection networks, it is a common practice to enlarge the input image, making the problem even worse. 
The second disadvantage is the lack of relative position encoding in the module.
Coupling both appearance and location information has been proven beneficial in several vision tasks including segmentation \cite{krahenbuhl2011efficient} and image restoration \cite{liu2018non} and is a core component in classical schemes that aggregate information from different locations in the image like bilateral filter \cite{tomasi1998bilateral} and various graphical models \cite{lafferty2001conditional, krahenbuhl2011efficient, chen2014semantic}.  


We modified the NL module to deal with the above difficulties. A simple modification, based on the associative law of matrix multiplication, and exploiting the existing factorization of 
the affinity matrix inside the module, enabled us to create a comparable building block with a linear complexity 
with respect to the spatial dimensions of the feature map. Relative position encoding was further added to the 
affinity matrix 
and gave the network the opportunity to use relative spatial information in an efficient manner. The resulting scheme still aggregates information across the entire image, but not uniformly. We named this module ENL: Efficient Non Local module.

In this paper, we use the ENL module as a reasoning module that passes information between related pixels, applying it sequentially along the top-down stream. Since it is applied also to high resolution feature maps, efficiently re-implementation of the module is essential. 
Unlike other approaches, which placed a relational module on the BU stream, or establish relations between already detected objects, our framework can apply pairwise reasoning in a coarse to fine manner, guiding the detection of small objects. Applying the relational module to finer layers, also enables the small objects themselves to exchange information between each other.

To summarize our contributions:

1. We redesigned the NL module in an efficient manner (ENL), improved it in terms of memory and number of operations, allowing it to be placed it anywhere along the network. 

2.	We incorporated relative spatial information into the ENL module reasoning process, in a novel approach that keeps the efficient design of the ENL. 

3.	We applied the new module sequentially to increasing resolution feature maps along the top-down stream, obtaining relational reasoning in a coarse-to-fine manner.

4. We show the effectiveness of our scheme, incorporating it into the Faster-RCNN \cite{ren2017faster} and Mask-RCNN \cite{he2017mask} pipelines
and improving state-of-the-art detection of small objects over the COCO dataset \cite{lin2014microsoft} by 1-2 AP points on various architectures. 

The improvements presented in this work go beyond the specific detection application: tasks including semantic segmentation, fine-grained localization, images restoration, image  generation processes, or other tasks in the image domain, which use an encoder-decoder framework and depend on fine image details are natural candidates for using the proposed framework.

\vspace*{-5mm}

\section{Related Work}
The current work combines two approaches used in the field of object detection: (a) modelling context through top down modulation and (b) using non local interactions in a deep learning framework. We briefly review related work in these domains. 

\vspace*{-3mm}

\paragraph{Bottom-Up Top-Down Networks}
In detection tasks, one of the major challenges is to detect simultaneously both large and small objects. Early works used for the task a pure bottom-up (BU) architecture, and predictions were made only from the coarsest (topmost) feature map \cite{girshick2014rich, girshick2015fast, redmon2016you, he2014spatial}.    
Later works, tried to exploit the inherent hierarchical structure of neural networks to create a multi-scale detection architecture. Some of these works performed detection using combined features from multiple layers  \cite{bell2016inside, hariharan2017object, kong2016hypernet}, while others performed detection in parallel from individual layers \cite{liu2016ssd, cai2016unified, liu2017receptive, shen2017dsod}. 

Recent methods incorporate context (from the last BU layer) with low level layers by adding skip connections in a bottom-up top-down (BUTD) architecture. Some schemes  \cite{shrivastava2016beyond, ronneberger2015u, newell2016stacked} used only the last layer of the top down (TD) network for prediction, while others \cite{lin2017feature, he2017mask, fu2017dssd, redmon2018yolov3} performed prediction from several layers along the TD stream.


The last described architecture supplies enhanced results, especially for small objects detection and was adopted in various detection schemes (e.g. one stage or two stages detection pipelines). It assumes to successfully incorporate multi scale BU data with semantic context from higher layers, serves as an elegant built-in context module. 

In the current work we further enhance the representation created in the layers along the TD stream, using the pair-wised information, supplied by the NL module, already shown to be complementary to the CNN information \cite{wang2017non}. We show that sequentially applying this complementary source of information, in a coarse to fine manner, helps detection, especially of small objects.

\vspace*{-7mm}

\paragraph{Modern Relational Reasoning}
Relational reasoning and messages passing between explicitly detected, or implicitly represented objects in the image, is an active and growing research area. Recent work in this area has been applied to scene understanding tasks (e.g. recognition \cite{chen2018iterative}, detection \cite{wang2017non, hu2017relation, raposo2017discovering}, segmentation \cite{yuan2018ocnet}) and for image generation tasks (GANs \cite{zhang2018self}, restoration \cite{liu2018non}).

For scene understanding tasks, two general approaches exist. The first approach can be called 'object-centric', as it models the relations between existing objects, previously detected by a detection framework \cite{hu2017relation}.  
In this case, a natural structure for formalizing relation is via Graph Neural Network \cite{gori2005new, scarselli2009graph, battaglia2018relational, bronstein2017geometric} 

The second approach applies relational interactions directly to CNN feature maps (in which objects are implicitly represented). In this case, a dedicated block (sometimes named non local module \cite{wang2017non}, relational module \cite{santoro2017simple} or self-attention module \cite{vaswani2017attention}) is integrated into the network without an additional supervision, in an end-to-end learnable manner. 
Due to its high memory demands, current schemes \cite{wang2017non, santoro2017simple, battaglia2016interaction, zhang2018self, yue2018compact}, even those that output a full-sized image \cite{zhang2018self, liu2018non}, apply the non-local module to the higher layers of the network \cite{zhang2018self}, or restrict their operation to patches of the original image \cite{liu2018non}. 
We, on the other hand, coupled the ENL module with spatial context and applied it along the top down stream to lower layers and spatially larger feature-maps, consequently enabling to integrate  the module anywhere along the network.



\vspace*{-2mm}

\section{Approach}
We will first briefly review the implementation details of the Non Local (NL) module then present our proposed efficient ENL module, specifying our modifications in detail. Performance analysis of the resulting ENL module is presented in section \ref{sec:perf}.         

\begin{figure*}
	\begin{center}
		\fbox{\includegraphics[width=11.2cm]{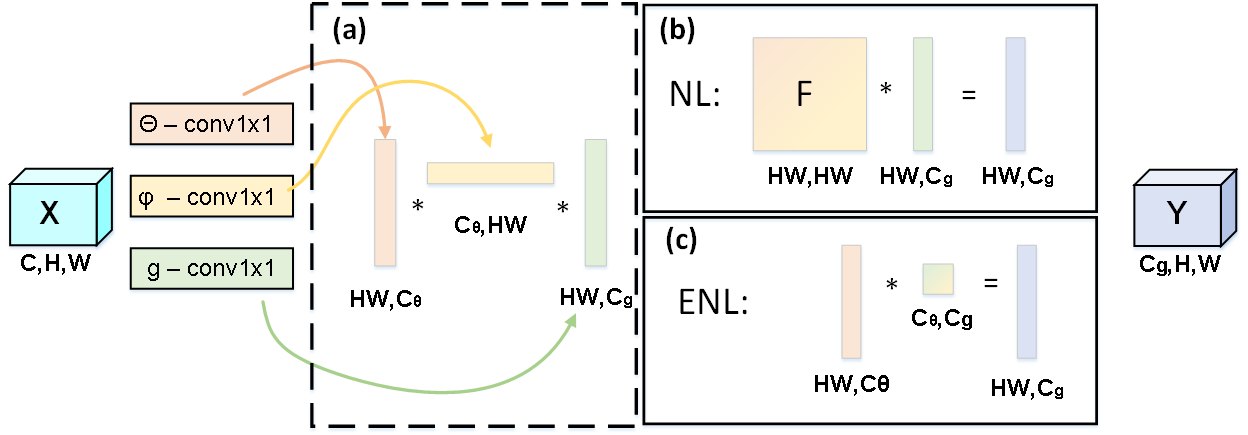}}
	\end{center}
	\caption{(b) NL module vs. (c) ENL module. Changing the order of matrices multiplication (a) decreases the number of operations and the memory utilization and practically allowing to use the non-local module anywhere along the network. }
	\label{fig:approach}
\end{figure*}

\vspace*{-7mm}

\subsection{Preliminaries: the Non Local module}
The formulation of the NL module as described in \cite{wang2017non} is:
\begin{equation}\label{eq:NLB}
y_i = \frac{1}{Z(x)}\sum_{j}F(x_i,x_j)g(x_j)
\end{equation}
Here $x$ is the input tensor and $y$ is the output tensor. 
Both $x$ and $y$ are flattened, namely modified to the form of $(HW \times \textit{channels})$ with H, W, the spatial dimensions of the feature map. 
$i \in [1, HW]$ is the current pixel under consideration, and $j$ runs over all spatial locations of the input tensor. The affinity matrix (also refered to as the similarity function) $\textbf{F} \in \mathbf{R}^{HW\times HW}$ summarizes the similarity between every two pixels $(i, j)$ in the input tensor ($F(x_i, x_j)$ is a scalar), Z(x) is a normalization factor and $g(x_j)$ is the representation of the $j$'th spatial pixel ($g\in \mathbf{R}^{HW\times C_g}$, $C_g$ channels in each pixel's representation). The module sums information from all the pixels in the input tensor weighted by their similarity to the $i$'th pixel. This equation can be alternatively written in matrix form (with the normalization factor merged into F):
\begin{equation}\label{eq:NLB_mat}
\textbf{y} = \textbf{F} \cdot \textbf{g}
\end{equation}

The similarity function $F(x_i,x_j)$, can be chosen in different ways; one of the popular design choices is:

\begin{equation}\label{eq:f_sim}
F(x_i, x_j)=e^{\theta(x_i)^T\phi(x_j)}
\end{equation}
Where $\theta$ and $\phi$ are linear transformations of $x$ that can be formulated as low rank matrices and implemented by a \textit{1x1conv} operation.
In this case the normalization factor $Z(x)$ takes the form of the \textit{softmax} operation. A block scheme of this implementation is illustrated in figure \ref{fig:approach}(a+b) with the related matrices sizes.

The described NL module goes through another \textit{1x1conv} and combined with a residual connection to take the form of:
\begin{equation}\label{eq:f_sim}
z_i=conv(y_i)+x_i
\end{equation}
For simplicity we omit the description of these operators from the rest of the section as well as from the illustration (figure \ref{fig:approach}). 

Two drawbacks of this basic implementation are its extremely expensive memory utilization and the lack of position encoding. Both of these issues are addressed next.

\vspace*{-3mm}

\subsection{ENL: Memory Effective Implementation}\label{sec:ENL}
Let us consider the case of another design choice of $F(x_i, x_j)$:
\begin{equation}\label{eq:f_no_exp}
F(x_i, x_j)=\theta(x_i)^T\phi(x_j)
\end{equation}

In this case $F$ is a $HW\times HW$ matrix created by a multiplication of two matrices, $\theta^T\in\mathbf{R}^{HW \times C_\theta}$ and $\phi\in\mathbf{R}^{C_\theta \times HW}$ (figure \ref{fig:approach}a). Since this matrix multiplication is immediately followed by another matrix multiplication with $g\in\mathbf{R}^{HW \times C_g}$ - one can simply use the associative rule to change the order of the calculation, written in matrix form as:

\begin{equation}\label{eq:mat_mul_1}
\textbf{y} = \textbf{F} \cdot \textbf{g} = 
\textbf{$\theta^T$} \cdot \textbf{$\phi$} \cdot \textbf{g} = 
\textbf{$\theta^T$} \cdot \left[ \textbf{$\phi$} \cdot \textbf{g} \right] 
\end{equation}

An illustration of the memory effective implementation with the corresponding matrices dimensions can be visualized in figure \ref{fig:approach}(a+c).


This re-ordering results in a large saving in terms of memory and operations used.  Consider a detection framework with typical image size of $800\times1000$. In the second stage (stride 4) $HW\approx 50,000$. While the inner multiplication result by sequentially multiplying the matrices (original NL, figure \ref{fig:approach}b) is $F\in\mathbf{R}^{HW\times HW}$, the multiplication reordering (ENL module \ref{fig:approach}c) gives an inner result of size $\in\mathbf{R}^{C_{\theta} \times C_g}$ (at least 4 orders reduction of memory utilization inside the block). The reduction in the number of operations is determined in a similar manner, see section \ref{sec:perf} for a detailed performance analysis. 

\vspace*{-3mm}

\subsection{Adding Relative Position Encoding}\label{sec:EBNL}

\begin{figure}
	\begin{tabular}{ccccc}
		\bmvaHangBox{\fbox{\includegraphics[width=2.0cm]{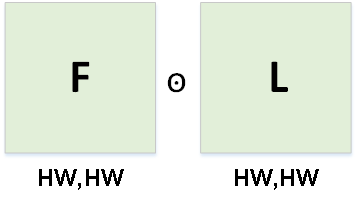}}}&
		\bmvaHangBox{\fbox{\includegraphics[width=1.8cm]{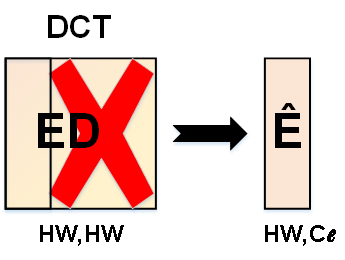}}}&
		\bmvaHangBox{\fbox{\includegraphics[width=2.4cm]{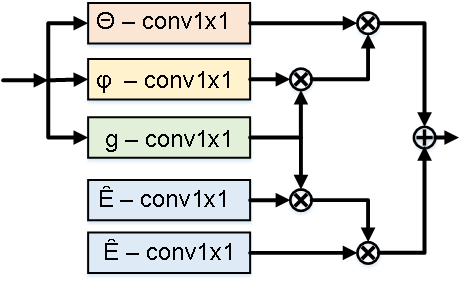}}}&
		\bmvaHangBox{\fbox{\includegraphics[width=1.6cm]{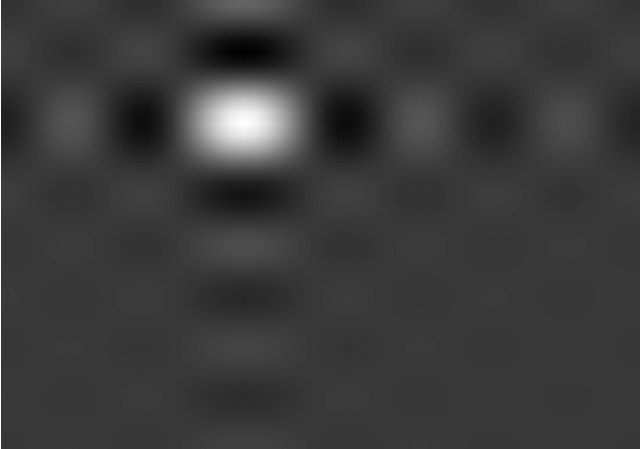}}}&
		\bmvaHangBox{\fbox{\includegraphics[width=1.6cm]{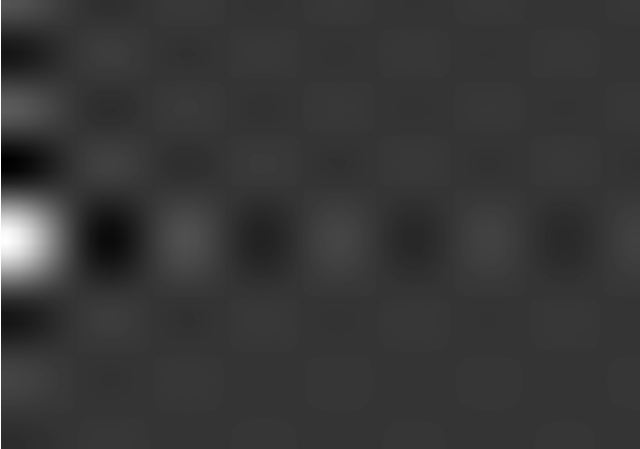}}}\\
		(a)&(b)&(c)&(d)&(e)
	\end{tabular}
	\vspace{0.3cm}
	\caption{(a) adding position encoding to the NL module is straightforward. (b,c) adding position encoding to the ENL module by constructing $\hat{E}$. (d,e) two examples of the resulting filters}
	\label{fig:NL_ENL_pos}
	\vspace*{-3mm}
\end{figure}

Adding a relative position encoding to the \textbf{original} non-local module is a straightforward procedure: Since $\textbf{F}\in\mathbf{R}^{HW\times HW}$ is explicitly calculated it can be element wise multiplied by (or added to) any function of the general form $L(i,j)$, $\textbf{L} \in\mathbf{R}^{HW\times HW}$. 

\begin{equation}\label{eq:full_spat}
\tilde{F}(x_i, x_j)=F(x_i,x_j)\odot L(i,j)
\end{equation}

With $\odot$ denotes an elementwise operation. See figure \ref{fig:NL_ENL_pos}a.

On the other hand, applying a relative position encoding to the \textbf{efficient} non-local module in a manner that will keep its low rank properties is a more  challenging task. 





In order to keep the low-rank properties of \textbf{L} we decompose it to take the form of:

\begin{equation}\label{eq:l_spat}
\textbf{L}=\hat{\textbf{E}}\cdot \hat{\textbf{E}}^T
\end{equation}

With $\hat{\textbf{E}} \in\mathbf{R}^{HW\times C_l}$, $C_l << HW$. Applying the associative rule and change the order of the matrices multiplication, in analogy to equation \ref{eq:mat_mul_1}, yields:

\vspace*{-5mm}

\begin{equation} \label{eq:cheap_spat_add}
\begin{split}
\textbf{y} = \tilde{\textbf{F}} \cdot \textbf{g} =  \left[ \textbf{F} + \textbf{L} \right]\cdot \textbf{g} &= \left[ \textbf{$\theta^T$}\cdot\textbf{$\phi$}+ \hat{\textbf{E}}\cdot \hat{\textbf{E}}^T \right] \cdot \textbf{g}  = \\
& = \textbf{$\theta^T$}\cdot \left[ \textbf{$\phi$} \cdot \textbf{g} \right] + \hat{\textbf{E}}\cdot  \left[ \hat{\textbf{E}}^T \cdot \textbf{g}  \right]
\end{split}
\end{equation}

The last equation can be easily used to build a block scheme of the proposed module (see figure \ref{fig:NL_ENL_pos}c). 

Next we will explicitly construct $\hat{\textbf{E}}$.
Denote $\textbf{E}\in \mathbf{R}^{HW\times HW}$ as an arbitrary real valued matrix and $\textbf{D}\in \mathbf{R}^{HW\times HW}$ as a diagonal matrix with only $C_l$ non-zero elements. As a result, $\textbf{E} \cdot \textbf{D}$ is a matrix with only $C_l$ non-zero columns. Notice that since $\textbf{D}$ is a diagonal matrix, we can easily find $\textbf{K}_D\in \mathbf{R}^{HW\times HW}$ that satisfy $\textbf{K}_D \odot \textbf{A} = \textbf{D} \cdot \textbf{A}$ for any given matrix $\textbf{A}$ where $\odot$ is an element-wise multiplication and $\cdot$ is a dot product. Denote $\hat{\textbf{E}}$ as the concatenation of the non-zero columns of $\textbf{E} \cdot \textbf{D}$ (see a schematic illustration in figure \ref{fig:NL_ENL_pos}b). By construction:

\vspace*{-2mm}

\begin{equation}\label{eq:l_spat}
\textbf{L}=\hat{\textbf{E}}\cdot \hat{\textbf{E}}^T
= \textbf{E} \cdot \textbf{D} \cdot \textbf{D}^T \cdot \textbf{E}^T
= \textbf{E} \cdot \left| \textbf{D} \right|^2 \cdot \textbf{E}^T
= \textbf{E} \cdot \left[ \textbf{K}_D \odot  \textbf{K}_D \odot \textbf{E}^T \right]
= \textbf{E} \cdot \left[ \textbf{K}_D^{\odot 2} \odot \textbf{E}^T \right]
\end{equation}

Where $\textbf{K}_D^{\odot 2} = \textbf{K}_D \odot  \textbf{K}_D$. 
$\hat{\textbf{E}} \hat{\textbf{E}}^T$ can therefore be used as an efficient way to calculate $\textbf{E} \textbf{D} \textbf{D}^T \textbf{E}^T$ for any arbitrary matrices $\textbf{E}$ and $\textbf{D}$.

We choose E as the matrix form of the 2D-DCT (each row of E consists of the 2D-DCT coefficients of an image with 1 in the corresponding single coordinate and 0 anywhere else, flattened to the form of $(1 \times HW)$). Notice that  $\textbf{E}^T \cdot g$ is the result of applying the DCT to a signal $g$ and $\textbf{E} \cdot \textbf{G}$ indicates applying the inverse transform on  $\textbf{G}$. Choosing $\textbf{E}$ as the 2D-DCT matrix meets several goals:

a. $\hat{\textbf{E}} \hat{\textbf{E}}^T$ can be interpreted as a spatial filter (see below).

b. Common DCT properties (the DCT of a multiplication equals the convolution of the DCTs, the DCT is a unitary matrix) can be used.

c. The DCT matrix concentrate most of the signal energy in a small number of coefficients that corresponds to its lower frequencies.

Using the convolution theorem on the result of equation \ref{eq:l_spat} yields:


\vspace*{-5mm}

\begin{equation}
\begin{array}{cccc}
& \scriptsize{\textrm{$DCT^{-1}$ of a multiplication}} &  \scriptsize{\textrm{convolution of $DCTs^{-1}$}} & \\
\textbf{L} \cdot g 
=\hat{\textbf{E}}\cdot \hat{\textbf{E}}^T \cdot g  &
= \overbrace{\textbf{E} \left[ \textbf{K}_D^{\odot 2} \odot \textbf{E}^T  g \right]}&
= \overbrace{\textbf{E} \textbf{K}_D^{\odot 2} \ast \underbrace{\textbf{E} \left[  \textbf{E}^T  g \right]}} = &  
\underbrace{\textbf{E} \left[ \textbf{K}_D \odot  \textbf{K}_D \right] } \ast  g \\
& & \hspace{1.0cm} \scriptsize{\textrm{E is a unitary matrix}} & \scriptsize{\textrm{the filter}} \\
\end{array}
\end{equation}

Where $\ast$ is a symmetric convolution operation \cite{martucci1994symmetric} and $\textbf{E}  \left[ \textbf{K}_D \odot  \textbf{K}_D \right]$ is the inverse 2D-DCT on $\left[ \textbf{K}_D \odot  \textbf{K}_D \right]$ that functions as a spatial filter.

We take the support of D (the non-zero entries on the diagonal of D) as the columns that correspond to the lower frequencies of the 2D-DCT. On our experiments, we used the $9 \times 9$ columns that correspond to the lowest frequencies and set all the non-zero elements in $\textbf{D}$ to one. Optimizing the choice of the columns or the weights on the main diagonal of $\textbf{D}$ can be done but is out of the scope of this paper. 
$\textbf{E} \left[ \textbf{K}_D \odot  \textbf{K}_D \right]$ can be geometrically interpreted as a sum of low frequencies cosine waves, resulting a sinc like filter. 
The resulting filter is spatial invariant up to edge effects; its general structure is kept although fluctuations in its height exist. Two arbitrary examples are demonstrated in figure \ref{fig:NL_ENL_pos}d,e. 

We combined $\hat{\textbf{E}}$ with the efficient implementation described in section \ref{sec:ENL} as proposed on equation \ref{eq:cheap_spat_add} and illustrated in figure \ref{fig:NL_ENL_pos}c, allowing our scheme to couple both appearance and location information in an efficient manner. We use this scheme in all of our experiments unless stated otherwise.  

\vspace*{-5mm}

\section{Experiments \& Results}\label{sec:exp}

We evaluated our framework on the task of object detection, comparing to Faster RCNN \cite{ren2017faster} and Mask RCNN \cite{he2017mask} as baselines and demonstrating a consistent improvement in performance. 

\subsection{Implementation details} 

We implemented our models on PyTorch using the maskrcnn-benchmark  \cite{massa2018mrcnn} with its standard running protocols. We performed our experiences using FPN \cite{lin2017feature} with several backbones, all pretrained on ImageNet \cite{deng2009imagenet}. We integrated three instances of the ENL modules sequentially along the top down stream. The images were normalized to 800 pixels on their shorter axis. All the models were trained on \textit{COCO train2017} (\cite{lin2014microsoft}, ~118K images) and were evaluated on \textit{COCO val2017} (5K images). We report our results based on the standard metrics of COCO, using \textit{AP} (mean average precision) and \textit{APsmall} (mean average precision for small objects, $\textit{area} < 32^2$) as our main criterions for comparison. Further imlementation details are specified in the Supplementary Materials.

\vspace*{-3mm}

\subsection{Comparison with state of the art results} 
Table \ref{tab:state_of_art} compares the detection results of the proposed scheme(\textbf{+3ENL, TD}, using Faster RCNN with three additional ENL modules along the TD stream) to the baseline (Faster RCNN) and to the variant suggested by \cite{wang2017non} (+1NL, BU). Adding efficient non local modules in a coarse to fine manner along the TD stream leads to a consistent improvement of up to 1.5 points in the detection rates of small objects (APsmall) over Faster RCNN and up to 1 point over adding a non local module on the BU stream.

Table \ref{tab:state_of_art_mask} shows 1-2 APsmall points improvement over Mask-RCNN, both for bbox prediction and for mask evaluation.


\begin{table}[ht]
	\begin{center}
		\footnotesize
		\begin{tabular}{|l|l|c|c|c|c|}
			\hline 
			& & AP & APs & APm & APl \\ 
			\hline\hline 
			& baseline & 36.71 & 21.11 & 39.85 & 48.14 \\
			R50 & +1NL, BU & 37.72 & 21.65 & 40.88 & 49.07 \\
			& \textbf{+3ENL, TD} & 37.75 & \textbf{22.54} & 41.41 & 48.73 \\  
			\hline\hline 
			& baseline & 39.11 & 22.98 & 42.35 & 50.50 \\
			R101 & +1NL, BU & 40.03 & 23.06 & 43.65 & 52.32 \\
			& \textbf{+3ENL, TD} & 39.85 & \textbf{23.96} & 43.30 & 51.87 \\  
			\hline\hline 
			& baseline & 41.23 & 25.11 & 45.11 & 52.89 \\
			X101 & +1NL, BU & 42.11 & 25.89 & 45.60 & 54.16 \\
			& \textbf{+3ENL, TD} & 41.76 & \textbf{26.29} & 45.96 & 53.04 \\  
			\hline  
		\end{tabular} 
	\end{center}
	\caption{\textbf{Faster RCNN results for various architectures}}
	\label{tab:state_of_art}
\end{table}

\begin{table}[ht]
	\begin{center}
		\footnotesize
		\begin{tabular}{|l|l|c|c|c|c|}
			\hline 
			& & \multicolumn{2}{c}{bbox} \vline & \multicolumn{2}{c}{mask} \vline \\
			\hline
			& & AP & APs & AP & APs \\ 
			\hline\hline 
			& baseline & 37.80 & 21.49 & 34.21 & 15.61 \\
			R50 & +1NL, BU & 38.42 & 22.33 & 34.86 & 16.04 \\
			& \textbf{+3ENL, TD} & 38.65 & \textbf{23.33} & 34.88 & \textbf{17.01} \\  
			\hline\hline 
			& baseline & 40.02 & 22.81 & 36.07 & 16.37 \\
			R101 & +1NL, BU & 40.85 & 23.91 & 36.90 & 17.51 \\
			& \textbf{+3ENL, TD} & 40.80 & \textbf{24.46} & 36.59 & \textbf{17.83} \\   
			\hline  
		\end{tabular} 
	\end{center}
	\caption{\textbf{Mask RCNN results on the COCO benchmark}} 
	\label{tab:state_of_art_mask}
\end{table}

\vspace*{-2mm}

\begin{figure*} [ht]
	\begin{center}
		\begin{subfigure}[b]{0.193\linewidth}
			\adjincludegraphics[height=4.6cm,trim={{.2\width} {.2\height} 0 0},clip,keepaspectratio]
			{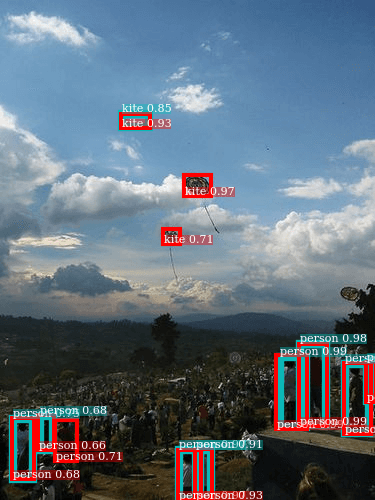}
			\caption{}
			\label{fig:successes_kites}
		\end{subfigure}
		\begin{subfigure}[b]{0.245\linewidth}
			\adjincludegraphics[height=4.6cm,trim={{.15\width} {.05\width} {.31\width} {.05\width}},clip,keepaspectratio]{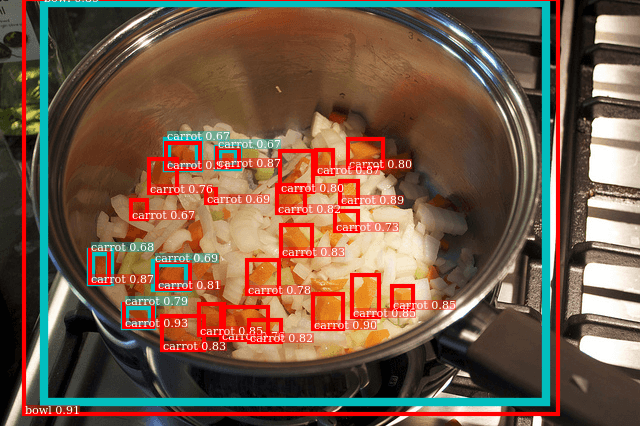}
			\caption{}
			\label{fig:successes_carrots}
		\end{subfigure}
		\begin{subfigure}[b]{0.12\linewidth}
			\adjincludegraphics[height=4.6cm,trim={{.2\width} 0 {.5\width} {.15\height}},clip,keepaspectratio]{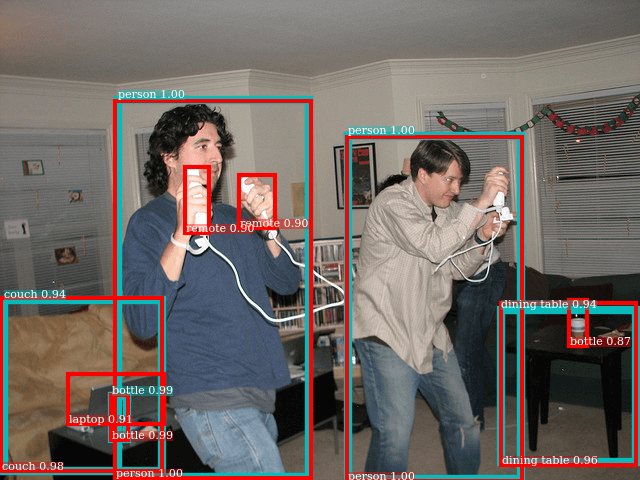}
			\caption{}
			\label{fig:successes_remotes}
		\end{subfigure}
		\begin{subfigure}[b]{0.135\linewidth}
			\adjincludegraphics[height=4.6cm,trim={{.2\width} 0 {.46\width} {.15\height}},clip,keepaspectratio]{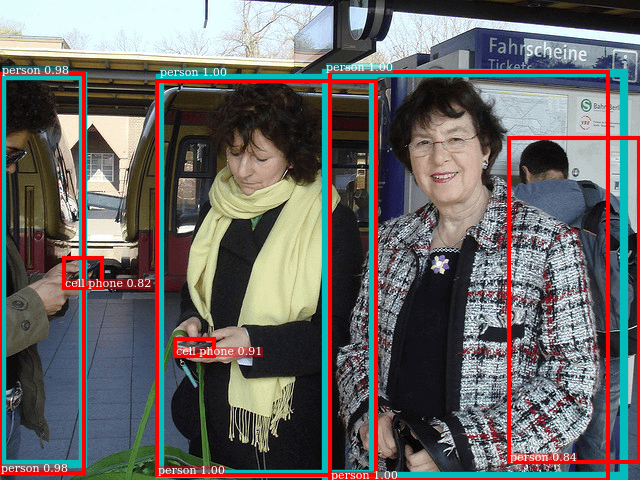}
			\caption{}
			\label{fig:successes_cellphone}
		\end{subfigure}
		\begin{subfigure}[b]{0.24\linewidth}
			\adjincludegraphics[height=4.6cm,trim={0 0 {.45\width} 0},clip,keepaspectratio]{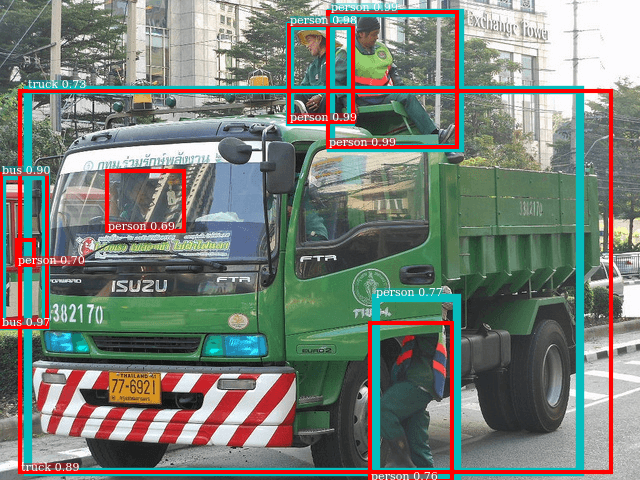}
			\caption{}
			\label{fig:successes_busdriver}
		\end{subfigure}
	\end{center}
	\caption{Qualitative examples from COCO val2017. (a-b). Repetitions of the same class in the image is exploited for detection. (c-e). Highly semantic clues in a top-down architecture: (c) The man is holding remotes in his hands, (d) The woman is watching her cellphone and (e) There is a driver in the truck. Cyan - NL module, Red - ours ENL. Same thresholds. Best if zoomed-in.}
	\label{fig:successes}
\end{figure*}

\paragraph {Qualitative results} 
Examples of interest are shown in Figure \ref{fig:successes}. The examples illustrate that small objects, that cannot be detected on their own, detected either by the presence of other instances of the same class (a,b) or by larger instances of related classes: (c) The man is holding two remotes in his hands, (d) the woman is holding and watching her cellphone and (e) detecting the driver inside the truck. These objects, marked in red, were not detected by Faster RCNN or by Faster RCNN with non-local module on the BU stream (using the same threshold). Please refer to the Supplementary Materials for additional qualitative examples.


\subsection{Performance analysis}\label{sec:perf}

Table \ref{tab:performance} shows the number of operation (Gflops), the memory utilization (MB, as measured on the GPU itself) and the number of params \textbf{in evaluation mode} with a ResNet50 backbone. Note that this analysis cannot be performed in training mode due to the extreme memory demands of the original NL module. The last two lines in the table emphasize the ENL efficiency and compares between the ENL (ours, bolded, including the relative position encoding) and the original NL module placed similarly along the top down stream (differences are highlighted in red). The high number of params (shown in green) in the second row is mainly because of the large descriptor size ($\left| C\right| =1024$) on the high layers of the network. Our ENL is placed on the top down stream ($\left| C\right| =256$) and uses much less additional parameters.

\begin{table}[ht]
	\begin{center}
		\footnotesize
		\begin{tabular}{|l|c|c|c|}
			\hline 
			& Gflops & memory & \#params \\ 
			\hline\hline 
			baseline & 90.20 & 1350 MB & 41.48 M \\
			+1NL, BU & 106.92 & 1358 MB & \textcolor{green}{43.58 M} \\
			\textbf{+3ENL, TD} & \textcolor{red}{\textbf{93.87}} & \textcolor{red}{\textbf{1426 MB}} & \textcolor{green}{\textbf{41.88 M}}\\
			+3NL, TD & \textcolor{red}{146.85} & \textcolor{red}{7476 MB} & 41.88 M\\  
			\hline 
		\end{tabular} 
	\end{center}
	\caption{\textbf{Performance Analysis. Evaluation mode.} Averaged on COCOval (5000 images), Performing the same analysis on training mode exeeds 16GB standard GPU memory.}
	\label{tab:performance}
\end{table}

\section{Conclusions}
\vspace*{-3mm}	
We examined the possible use of several non local modules, arranged hierarchically along the top down stream to exploit the effects of context and relations among objects. We compared our method with the previous use of a non local module placed on the bottom-up network, and show 1 AP improvement in small objects detection. We suggest that this improvement is enabled by the coarse-to-fine use of pair-wise location information and show visual evidence in support of this possibility.  

In practice, applying the non local module to large feature maps is a memory demanding operation. We deal with this difficulty and introduced ENL - an attractive alternative to the Non Local block, which is efficient in terms of memory and operations, and which integrates the use of relative spatial information. The ENL allows the use of non local module in a general encoder-decoder framework and consequently, might contribute in future work to a wide range of applications (segmentation, images generation etc.).




\pagebreak

\rule{0pt}{1ex}

\begin{raggedright}
	
	\LARGE \bfseries\sffamily\textcolor{sec_title_color}{Efficient Coarse-to-Fine Non-Local Module for the	Detection of Small Objects: Supplementary Material }  \par
	\vskip 1.5em
	
\end{raggedright}

\begin{minipage}[t]{0.48\textwidth}
	
	\large Hila Levi \\
	\textcolor{sec_title_color}{hila.levi@weizmann.ac.il} \\
	[4pt]
	\large Shimon Ullman \\
	\textcolor{sec_title_color}{shimon.ullman@weizmann.ac.il}
	
\end{minipage}%
\begin{minipage}[t]{0.48\textwidth}
	\large Dept. of Computer Science and Applied Mathematics,
	The Weizmann Institute of Science, Israel
\end{minipage}

\vspace*{0.3cm} 
\par\noindent\rule{\textwidth}{0.4pt}

\setcounter{equation}{0}
\setcounter{figure}{0}
\setcounter{table}{0}
\setcounter{page}{1}
\setcounter{section}{0}

\vspace*{-5mm}

\section{Implementation Details}\label{sec:implementation}
We performed our experiences on Faster R-CNN \cite{ren2017faster} detection framework and Mask R-CNN \cite{he2017mask}, using FPN \cite{lin2017feature} with several backbones, all pretrained on ImageNet \cite{deng2009imagenet}. We implemented our models on PyTorch using the maskrcnn-benchmark  \cite{massa2018mrcnn}. We used the standard running protocols of Faster R-CNN and adjust our learning rates as suggested by \cite{goyal2017accurate}. The images were normalized to 800 pixels on their shorter axis. All the models were trained on \textit{COCO train2017} (\cite{lin2014microsoft}, ~118K images) and were evaluated on \textit{COCO val2017} (5K images).

\paragraph{Training}
We trained our models for 360000 iterations using a base learning rate of 0.005 and reducing it by a factor of 10 after 240000 and 320000 iterations. 
We used SGD optimization with momentum 0.9 and a weight decay of 0.0001. We froze the BN layers in the backbone and replaced them with an affine operation as the common strategy when fine-tuning with a small number of images per GPU.

\paragraph{Inference}
During inference we followed the common practice of \cite{ren2017faster, he2017mask}. 
We report our results based on the standard metrics of COCO, using \textit{AP} (mean average precision) and \textit{APsmall} (mean average precision for small objects, $\textit{area} < 32^2$) as our main criterions for comparison. Further explanations of the metrics can be found in \cite{coco_det_eval}.  

\paragraph{Non Local Block} 
We placed the NL modules along the top down stream. We used three instances of the NL module in total, and located them in each stage, just before the spatial interpolation (in parallel to \textit{res5}, \textit{res4} and \textit{res3} layers).  

We initialized the blocks weights with random Gaussian weights, $\sigma = 0.01$. We did not use additional BN layers inside the NL module (due to the relatively small minibatch), or affine layers (since no initialization is available).

\section{Qualitative results} 

We integrated the non-local module along the top-down stream, on the network's backbone, preceding to the region proposal network and the classifier and supply additional information for both recognition and localization tasks. We demonstrate the effect of this additional source of information on the RPN and on the detector through the following examples.    

\subsection{Region Proposals}

\begin{figure}
	\begin{center}
		\begin{subfigure}[b]{0.49\linewidth}
			\adjincludegraphics[width=6cm,trim={0 0 0 {.085\height}},clip,keepaspectratio]
			{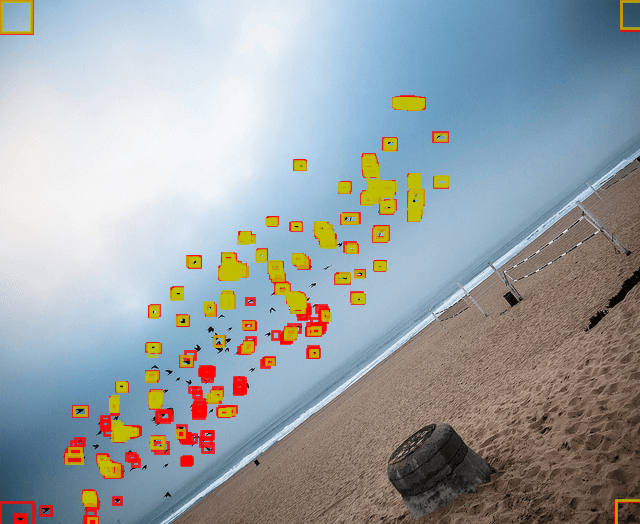}
			\caption{}
			\label{fig:rois_birds}
		\end{subfigure}
		\begin{subfigure}[b]{0.49\linewidth}
			\adjincludegraphics[width=6cm,trim={{.6\width} {.6\height} 0 0},clip,keepaspectratio]{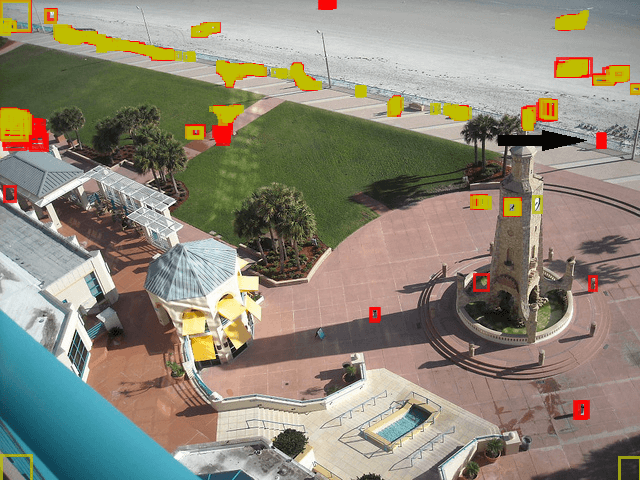}
			\caption{}
			\label{fig:rois_1man}
		\end{subfigure}
	\end{center}
	\caption{Using a reasoning module directly on the feature maps before the RPN enables more RoIs to pass to the next detection stage: (a) More than 10 birds were added to the flock. (b) A man watching on the sea. Yellow - Faster RCNN. Red (printed behind) - Ours. 1000 First RoIs are presented, for visually purposes only boxes smaller than 32x32 were printed.}
	\label{fig:rois_baseline_ver2}
\end{figure}

Using a reasoning module directly on the feature maps before the RPN potentially enables more RoIs to pass to the next detection stage. Examples, although rare (since this stage was designed for redundancy), exist. Figure \ref{fig:rois_baseline_ver2} shows examples of region proposals that wouldn't been passed to the detection stage unless applying the non-local module.

\subsection{Detections}

\begin{figure*}
	\setlength{\tabcolsep}{2pt}
	\begin{tabular}{cccc}
		\adjincludegraphics[width=3.1cm,trim={0 {.1\height} {.1\width} {.2\height}},clip,keepaspectratio]{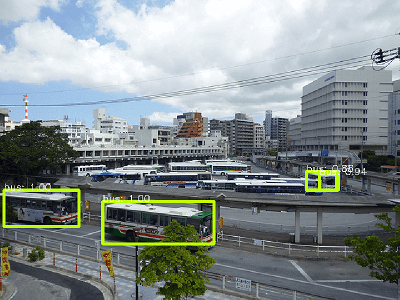}&
		\adjincludegraphics[width=3.1cm,trim={0 {.1\height} {.1\width} {.2\height}},clip,keepaspectratio]{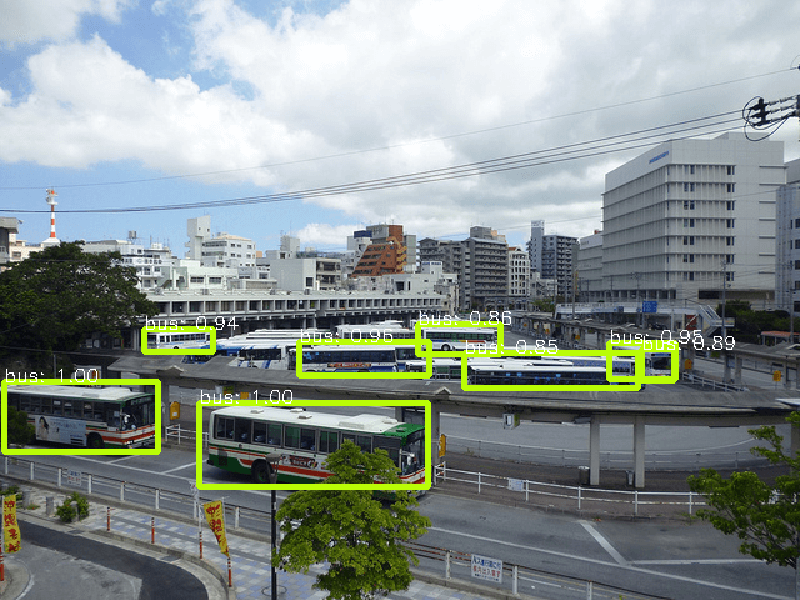}&
		\adjincludegraphics[width=3.1cm,trim={{.1\width} {.3\height} 0 0},clip,keepaspectratio]{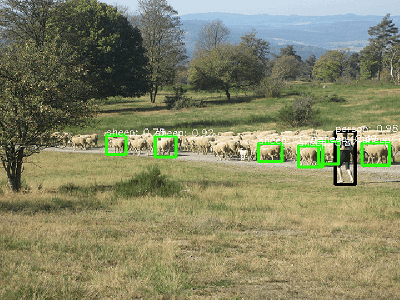}&
		\adjincludegraphics[width=3.1cm,trim={{.1\width} {.3\height} 0 0},clip,keepaspectratio]{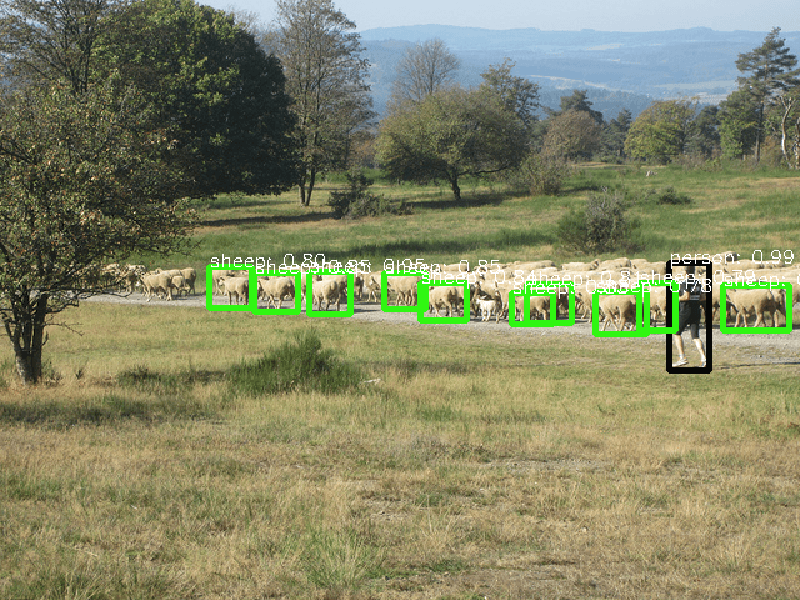}\\
		\adjincludegraphics[width=3.1cm,trim={0 0 0 {.27\height}},clip,keepaspectratio]{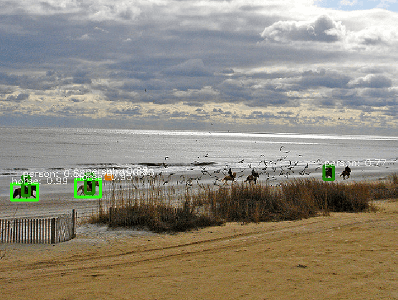}&
		\adjincludegraphics[width=3.1cm,trim={0 0 0 {.27\height}},clip,keepaspectratio]{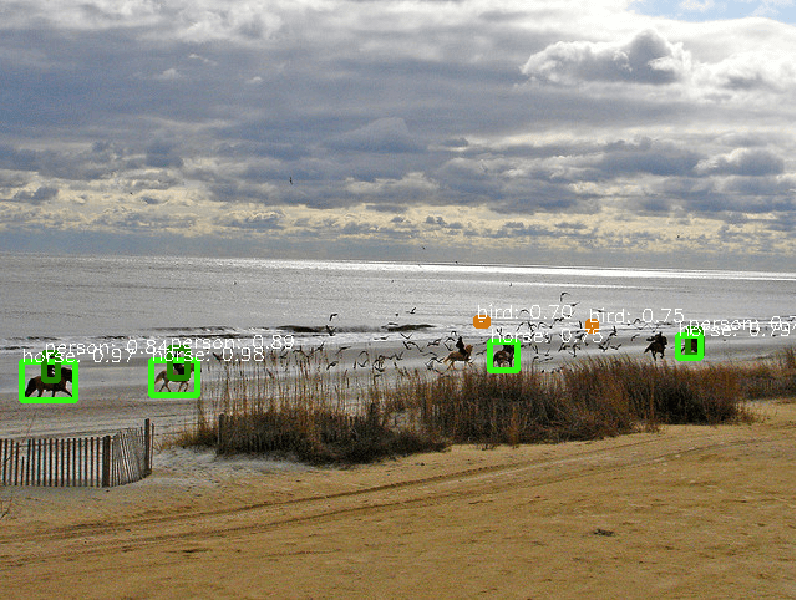}&
		\adjincludegraphics[width=3.1cm,trim={{.15\width} {.1\height} 0 {.2\height}},clip,keepaspectratio]{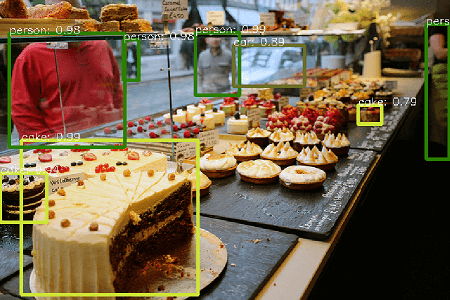}&
		\adjincludegraphics[width=3.1cm,trim={{.15\width} {.1\height} 0 {.2\height}},clip,keepaspectratio]{sm_pred_2252_TD.png}\\
		\adjincludegraphics[width=3.1cm,trim={0 {.1\height} 0 {.15\height}},clip,keepaspectratio]{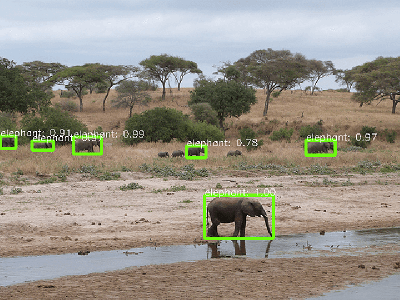}&
		\adjincludegraphics[width=3.1cm,trim={0 {.1\height} 0 {.15\height}},clip,keepaspectratio]{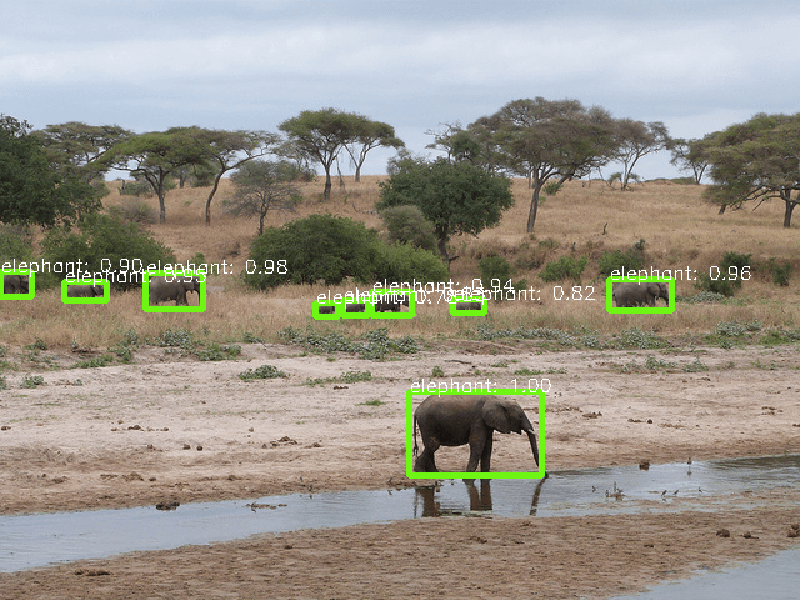}&
		\adjincludegraphics[width=3.1cm,trim={0 0 0 0},clip,keepaspectratio]{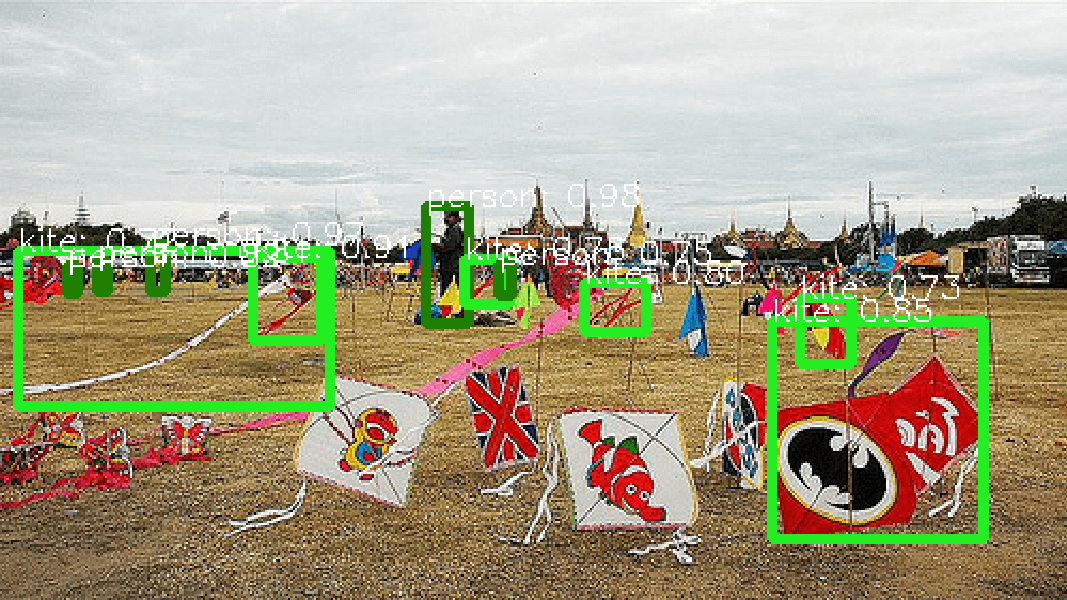}&
		\adjincludegraphics[width=3.1cm,trim={0 0 0 0},clip,keepaspectratio]{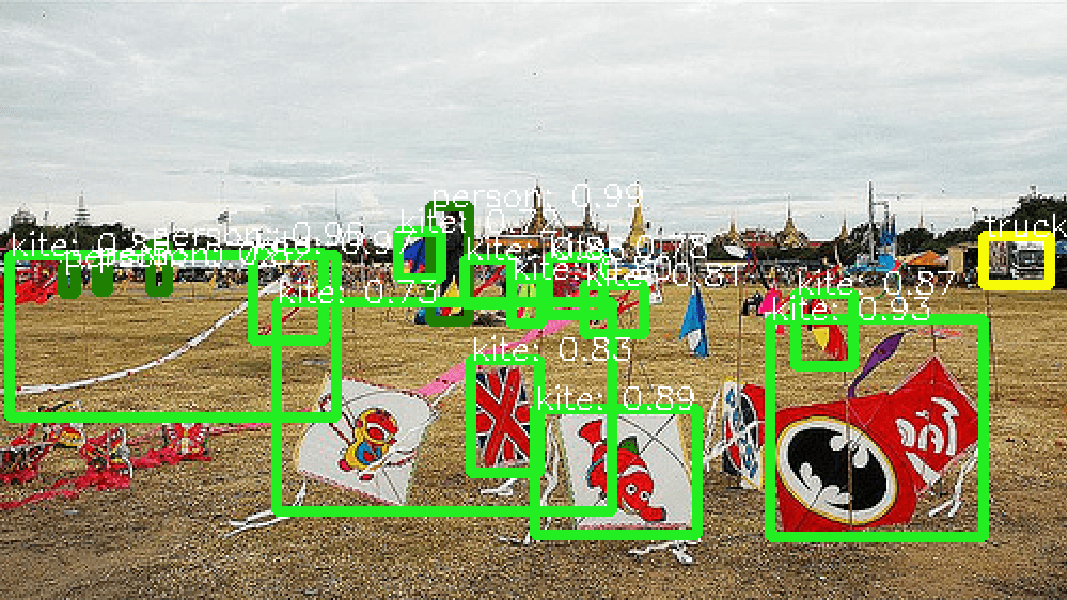}\\
		\adjincludegraphics[width=3.1cm,trim={0 {.25\height} {.3\width}  {.1\height}},clip,keepaspectratio]{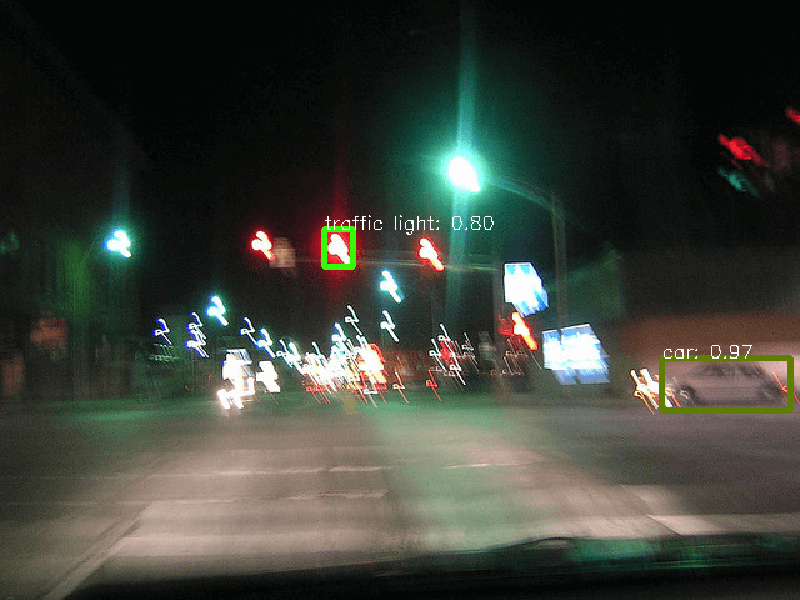}&
		\adjincludegraphics[width=3.1cm,trim={0 {.25\height} {.3\width}  {.1\height}},clip,keepaspectratio]{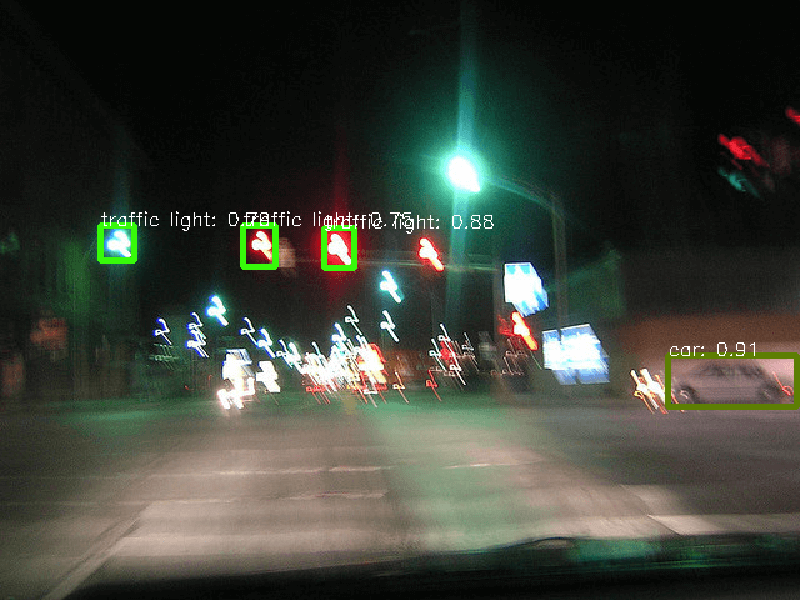}&
		\adjincludegraphics[width=3.1cm,trim={{.2\width} 0 0 {.18\height}},clip,keepaspectratio]{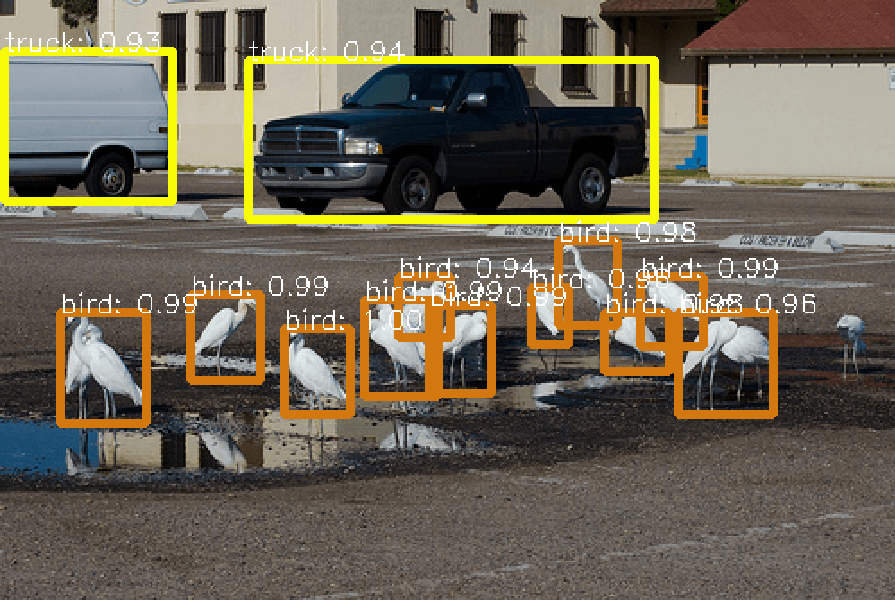}&
		\adjincludegraphics[width=3.1cm,trim={{.2\width} 0 0 {.18\height}},clip,keepaspectratio]{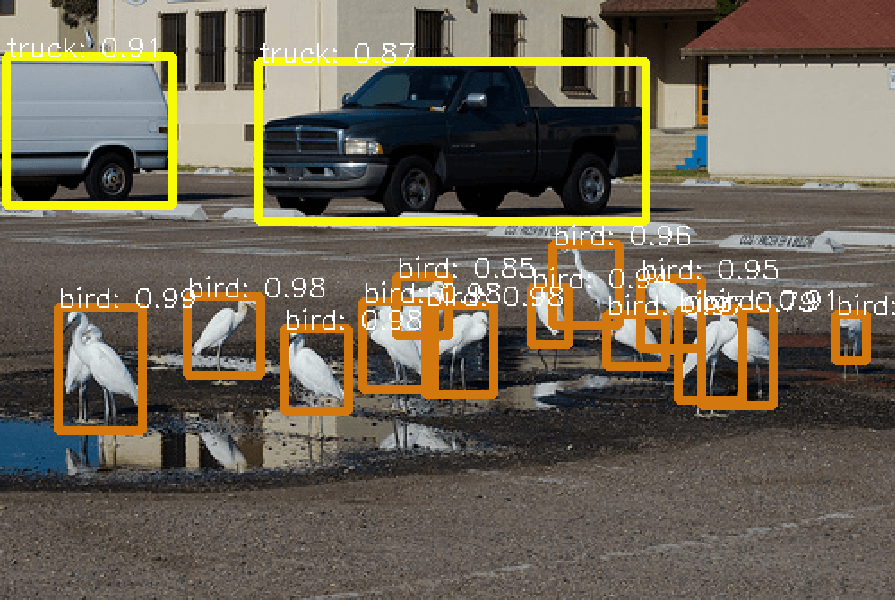}\\
		\adjincludegraphics[width=3.1cm,trim={0 {.25\height} 0 {.25\height}},clip,keepaspectratio]{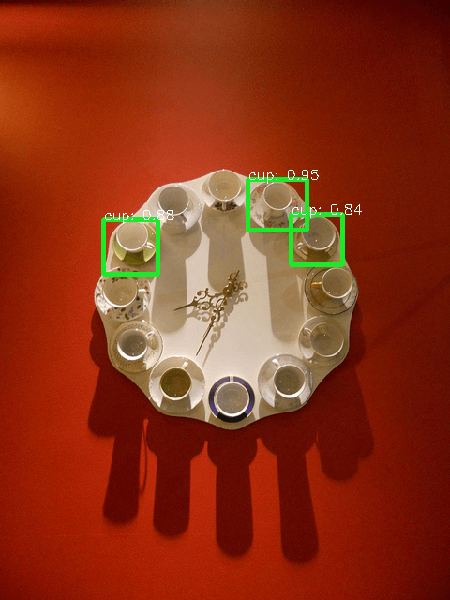}&
		\adjincludegraphics[width=3.1cm,trim={0 {.25\height} 0 {.25\height}},clip,keepaspectratio]{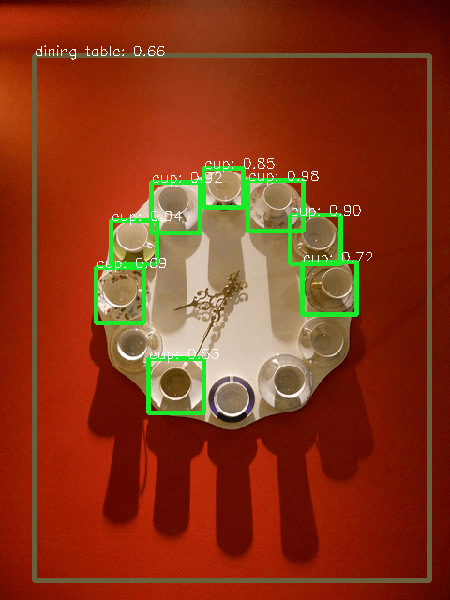}&
		\adjincludegraphics[width=3.1cm,trim={{.04\width} 0 {.18\width} 0},clip,keepaspectratio]{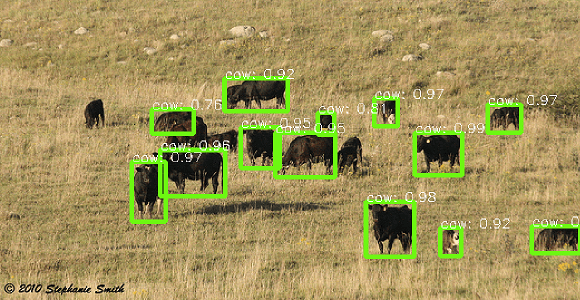}&
		\adjincludegraphics[width=3.1cm,trim={{.04\width} 0 {.18\width} 0},clip,keepaspectratio]{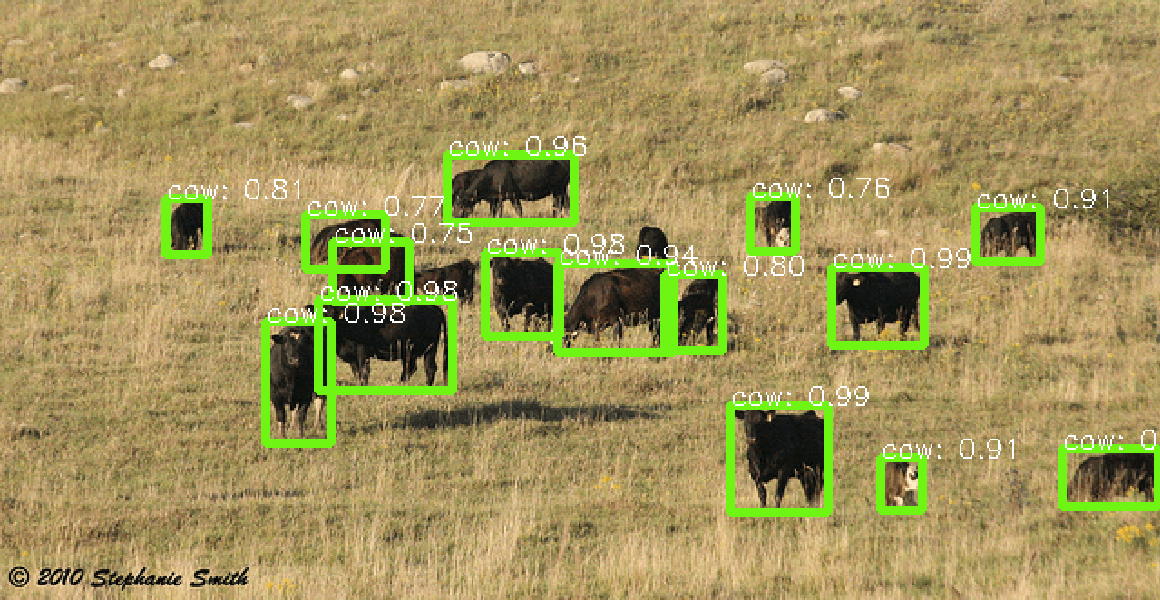}\\
		\adjincludegraphics[width=3.1cm,trim={0 0 0 0},clip,keepaspectratio]{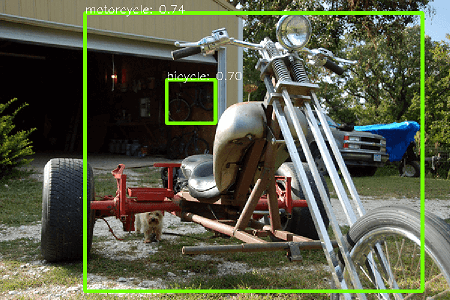}&
		\adjincludegraphics[width=3.1cm,trim={0 0 0 0},clip,keepaspectratio]{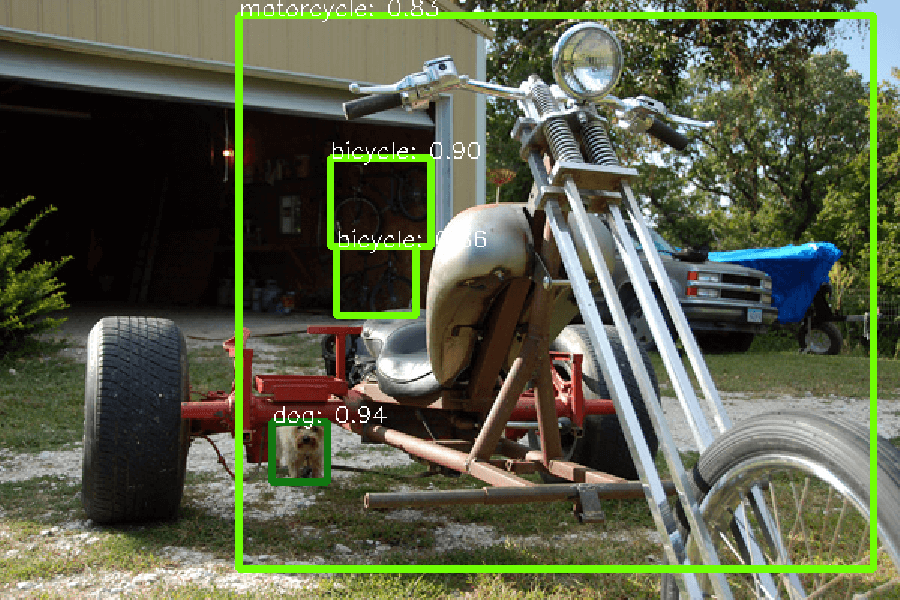}&
		\adjincludegraphics[width=3.1cm,trim={0 {.2\height} {.3\width} {.18\height}},clip,keepaspectratio]{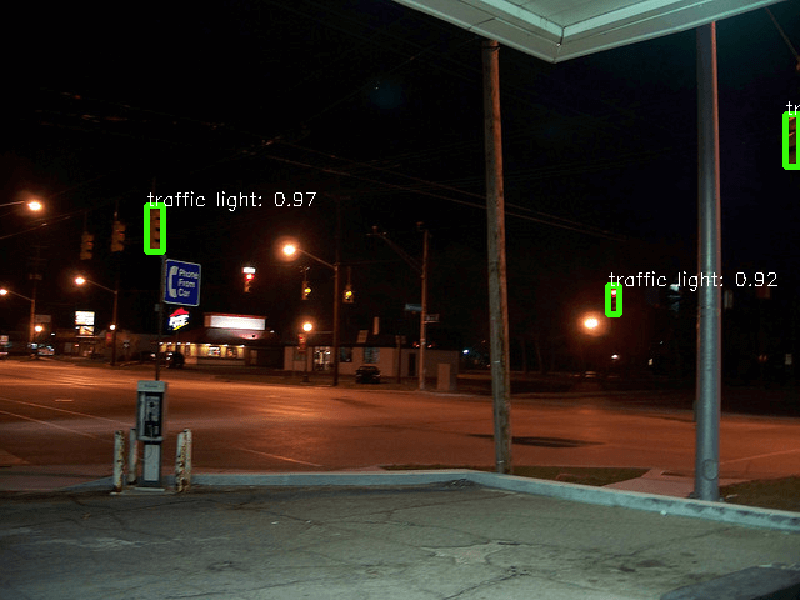}&
		\adjincludegraphics[width=3.1cm,trim={0 {.2\height} {.3\width} {.18\height}},clip,keepaspectratio]{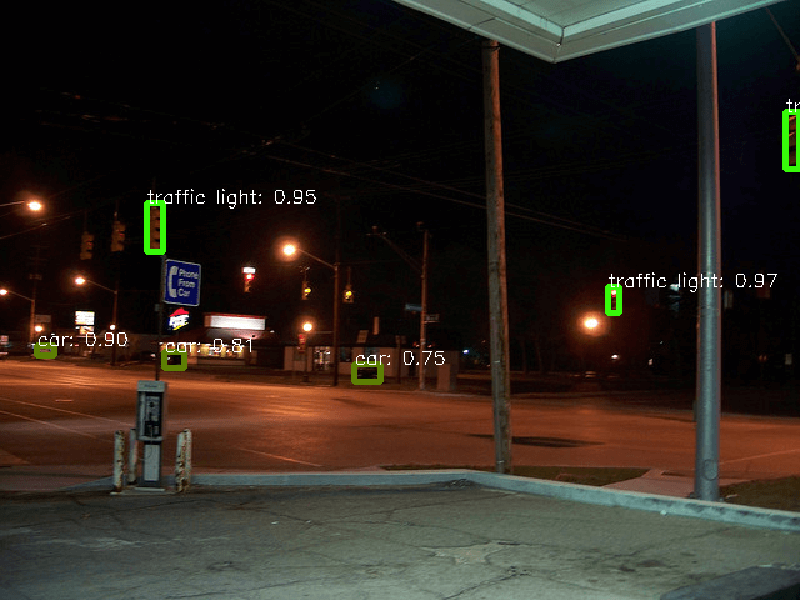}\\
		1NL, BU & 3ENL, TD (ours) & 1NL, BU & 3ENL, TD (ours) \\
		
	\end{tabular}
	\vspace{0.2cm}
	\caption{\textbf{Small objects, dense scenes:} ENL exploits repeatitions of small objects to yield better performance. Compared to \textit{1NL, BU} (non local module on the higher layers of the bottom up network, first and third columns), \textbf{3ENL, TD (ours)} (includes efficient non local modules along the top-down stream, second and forth columns) is capable to detect more instances and to better distiguish between overlapping instances. The last row demonstrates detections in a very dark scene. Best viewed while zoomed in. }
	\label{fig:small_objects}
\end{figure*}

\begin{figure*}
	\setlength{\tabcolsep}{2pt}
	\begin{tabular}{cccc}
		\adjincludegraphics[width=3.1cm,trim={0 {.3\height} 0 {.1\height}},clip,keepaspectratio]{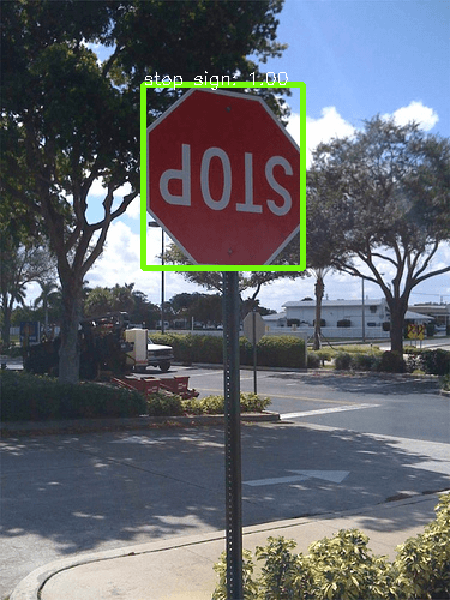}&
		\adjincludegraphics[width=3.1cm,trim={0 {.3\height} 0 {.1\height}},clip,keepaspectratio]{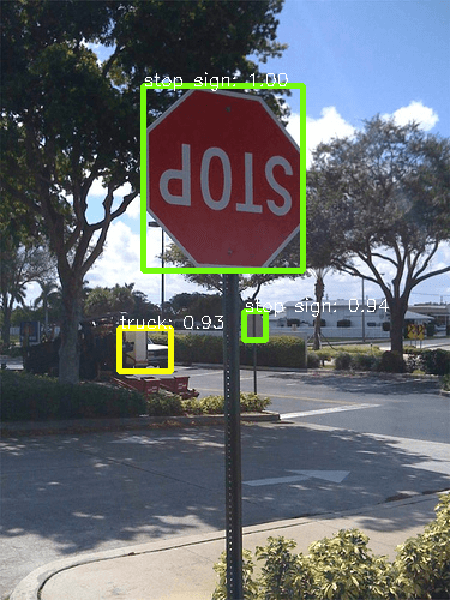}&
		\adjincludegraphics[width=3.1cm,trim={0 {.4\height} 0 0},clip,keepaspectratio]{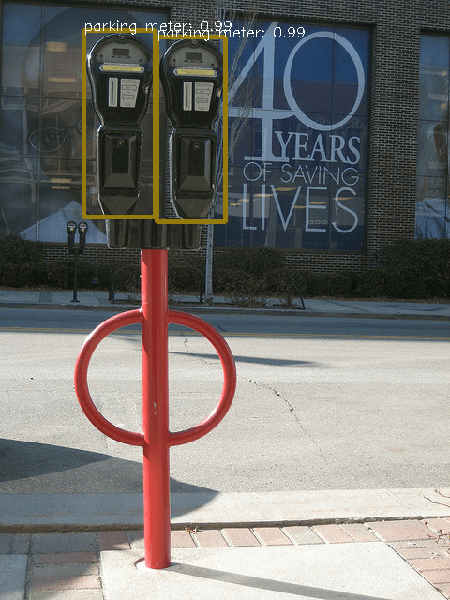}&
		\adjincludegraphics[width=3.1cm,trim={0 {.4\height} 0 0},clip,keepaspectratio]{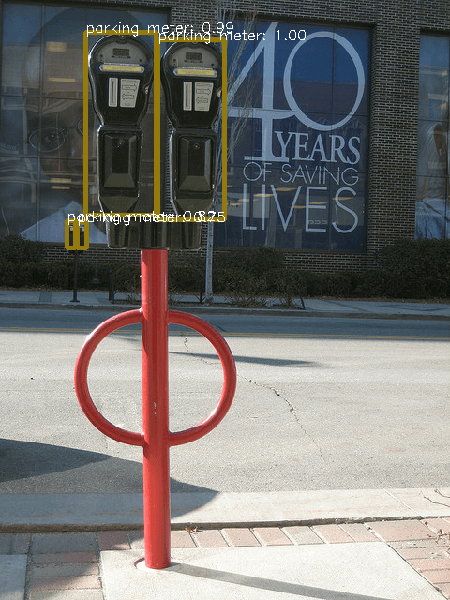}\\
		\adjincludegraphics[width=3.1cm,trim={0 0 0 {.1\height}},clip,keepaspectratio]{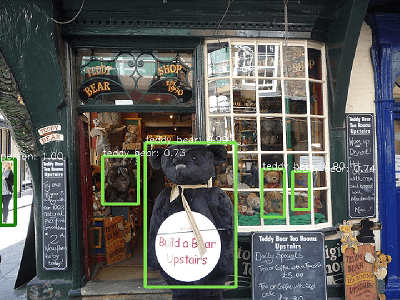}&
		\adjincludegraphics[width=3.1cm,trim={0 0 0 {.1\height}},clip,keepaspectratio]{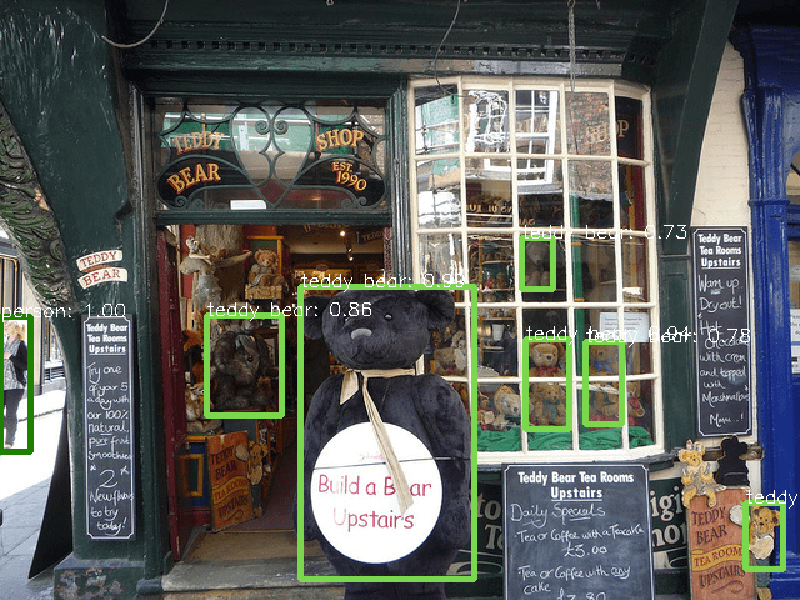}&
		\adjincludegraphics[width=3.1cm,trim={0 0 0 0},clip,keepaspectratio]{sm_pred_2896_BU.png}&
		\adjincludegraphics[width=3.1cm,trim={0 0 0 0},clip,keepaspectratio]{sm_pred_2896_TD.png}\\
		1NL, BU & 3ENL, TD (ours) & 1NL, BU & 3ENL, TD (ours) \\		
	\end{tabular}
	\vspace{0.2cm}
	\caption{\textbf{Top down modulation:} using ENL in a coarse-to-fine manner enables the detection of small instances partly based on the existence of larger instances of the same class. Best viewed while zoomed in.}
	\label{fig:top_down}
\end{figure*}

\begin{figure*}
	\setlength{\tabcolsep}{2pt}
	\begin{tabular}{cccc}
		\adjincludegraphics[width=3.1cm,trim={0 0 {.2\width} {.2\height}},clip,keepaspectratio]{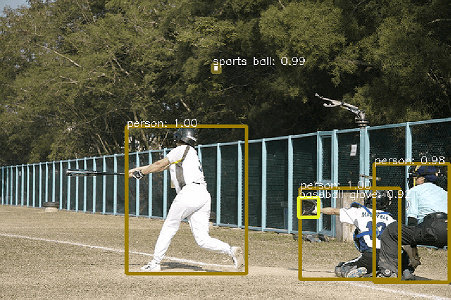}&
		\adjincludegraphics[width=3.1cm,trim={0 0 {.2\width} {.2\height}},clip,keepaspectratio]{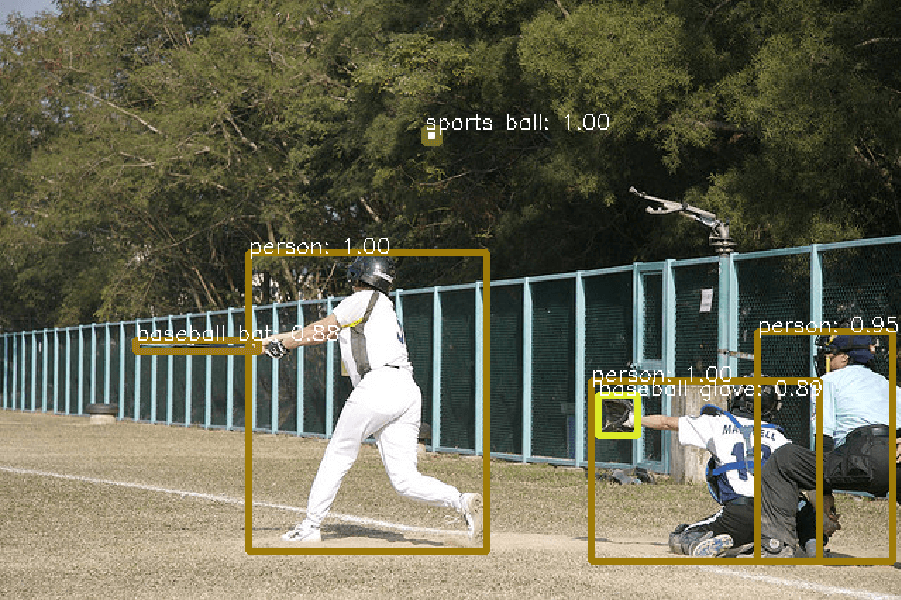}&
		\adjincludegraphics[width=3.1cm,trim={0 0 0 0},clip,keepaspectratio]{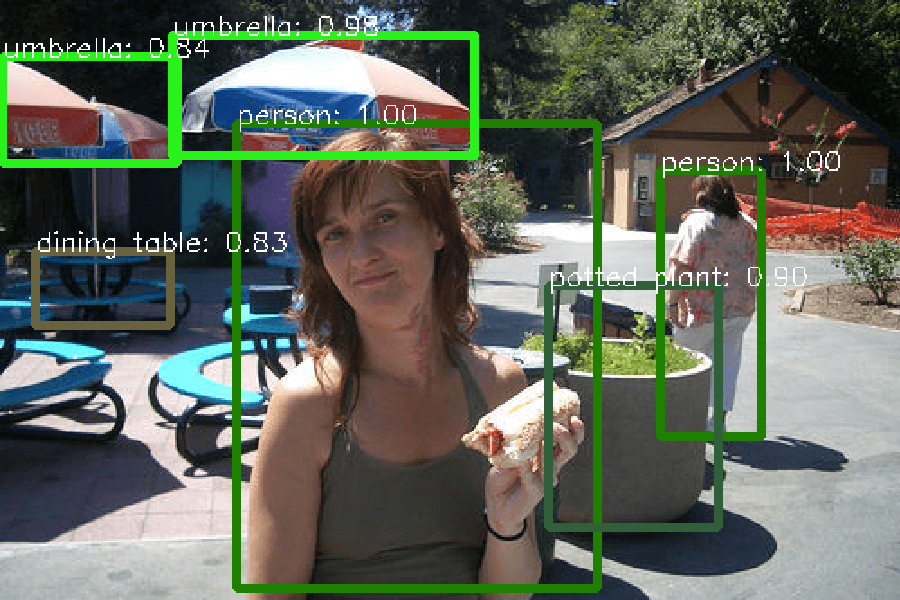}&
		\adjincludegraphics[width=3.1cm,trim={0 0 0 0},clip,keepaspectratio]{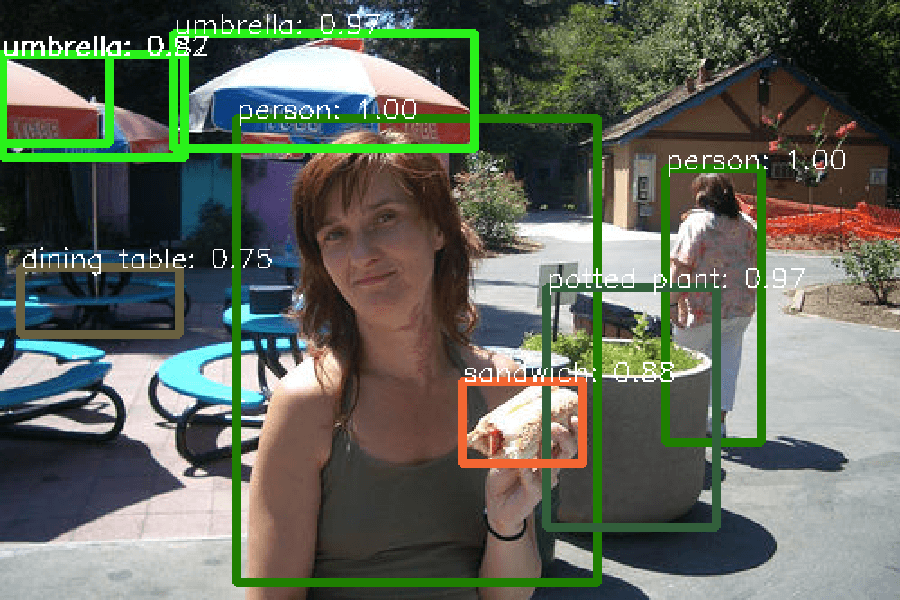}\\
		\adjincludegraphics[width=3.1cm,trim={0 {.1\height} 0 0},clip,keepaspectratio]{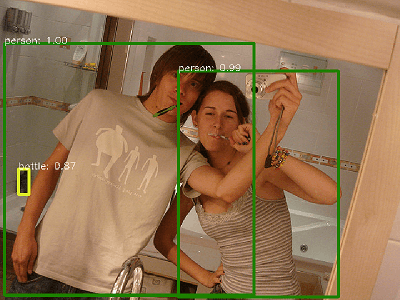}&
		\adjincludegraphics[width=3.1cm,trim={0 {.1\height} 0 0},clip,keepaspectratio]{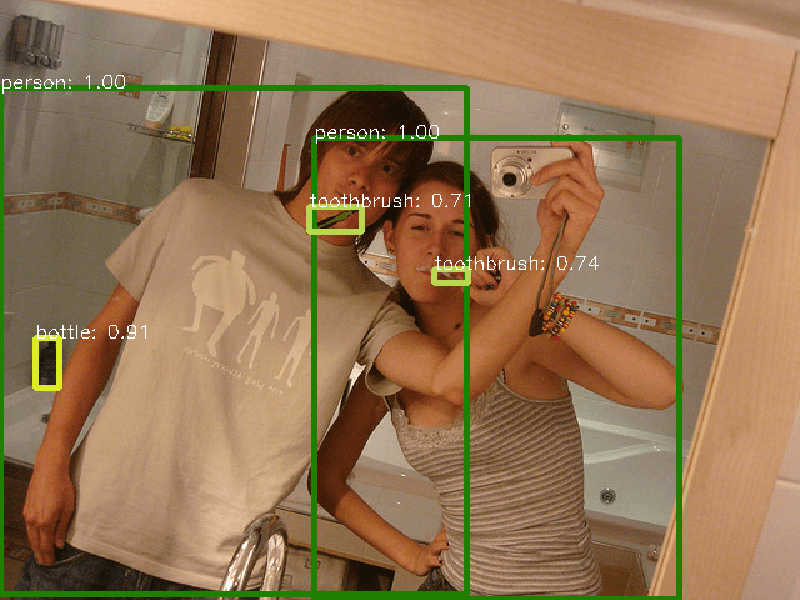}&
		\adjincludegraphics[width=3.1cm,trim={{.3\width} {.25\height} 0 {.15\height}},clip,keepaspectratio]{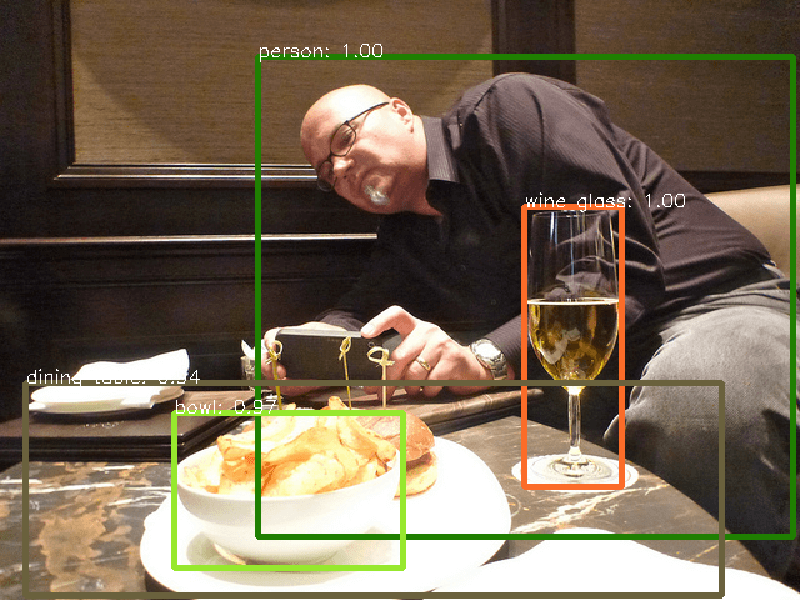}&
		\adjincludegraphics[width=3.1cm,trim={{.3\width} {.25\height} 0 {.15\height}},clip,keepaspectratio]{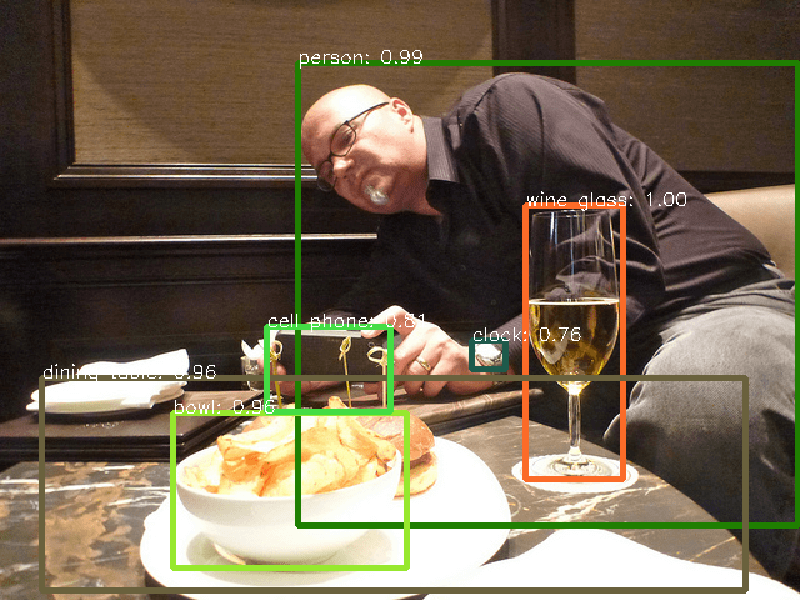}\\	
		\adjincludegraphics[width=3.1cm,trim={0 {.1\height} {.2\width} {.22\height}},clip,keepaspectratio]{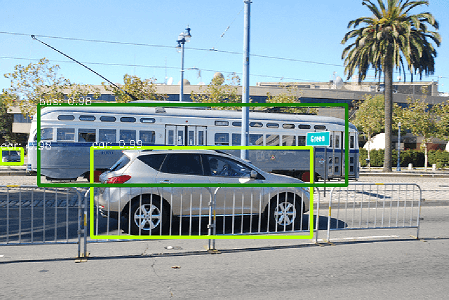}&
		\adjincludegraphics[width=3.1cm,trim={0 {.1\height} {.2\width} {.22\height}},clip,keepaspectratio]{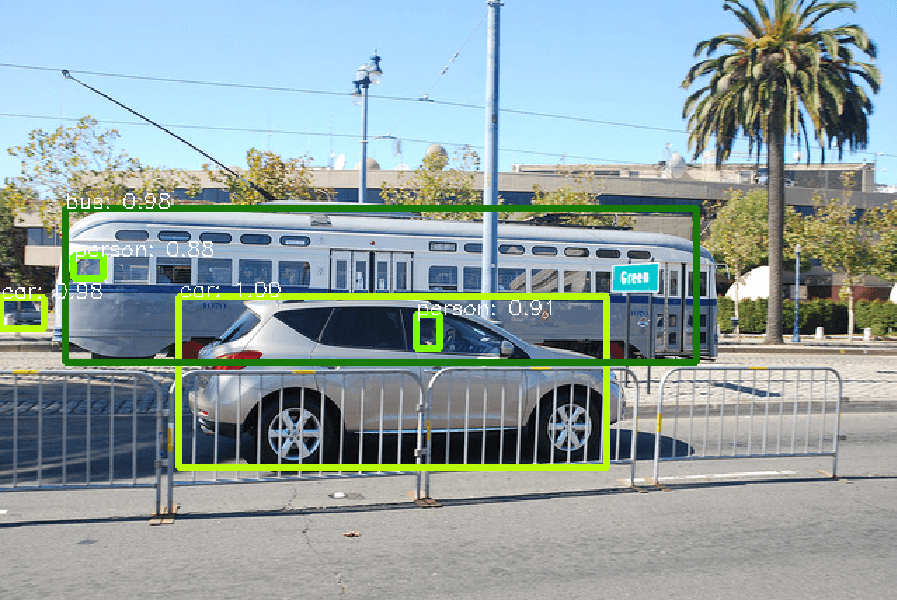}&
		\adjincludegraphics[width=3.1cm,trim={{.3\width} {.3\height} {.3\width} {.3\height}},clip,keepaspectratio]{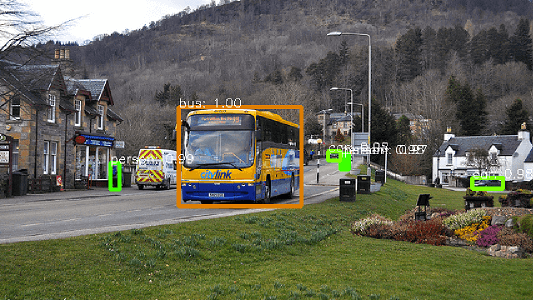}&
		\adjincludegraphics[width=3.1cm,trim={{.3\width} {.3\height} {.3\width} {.3\height}},clip,keepaspectratio]{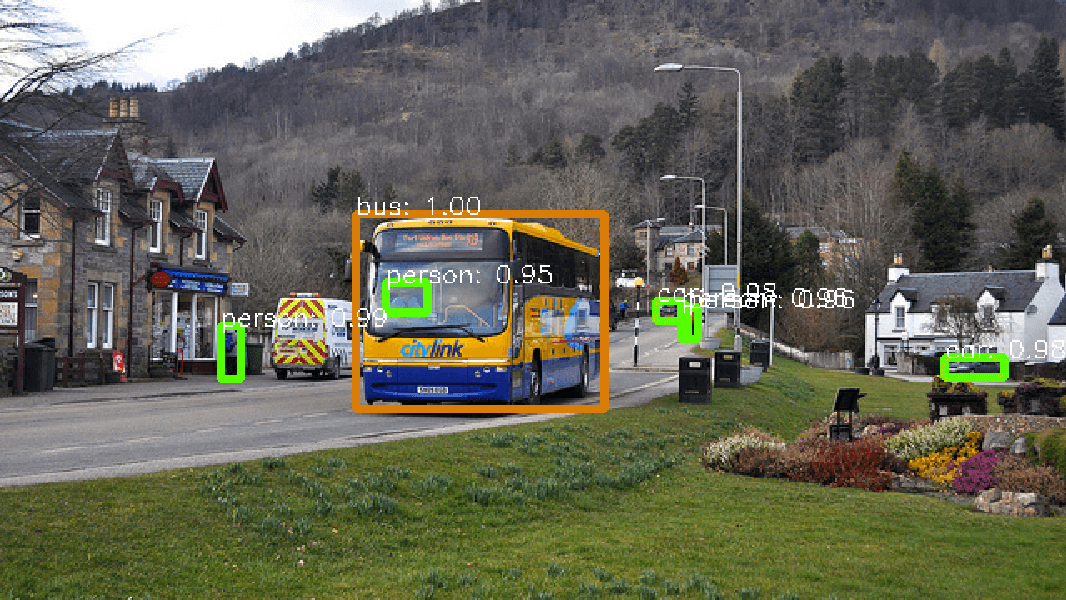}\\
		\adjincludegraphics[width=3.1cm,trim={0 {.3\height} {.4\width} 0},clip,keepaspectratio]{sm_pred_1457_BU.png}&
		\adjincludegraphics[width=3.1cm,trim={0 {.3\height} {.4\width} 0},clip,keepaspectratio]{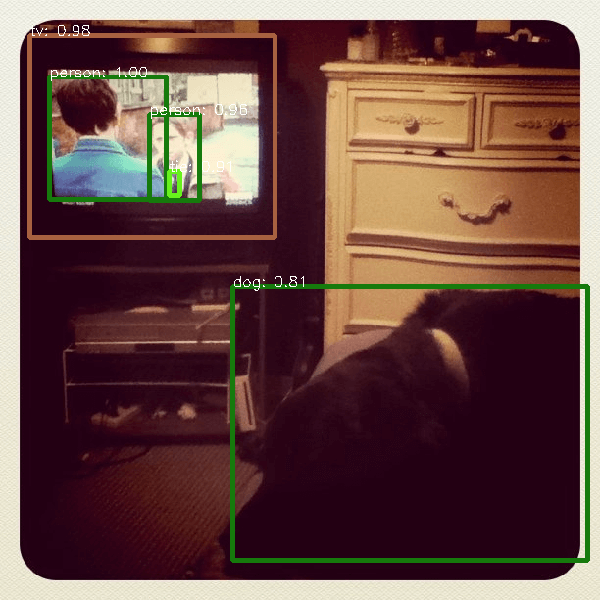}&
		\adjincludegraphics[width=3.1cm,trim={0 {.05\height} 0 {.1\height}},clip,keepaspectratio]{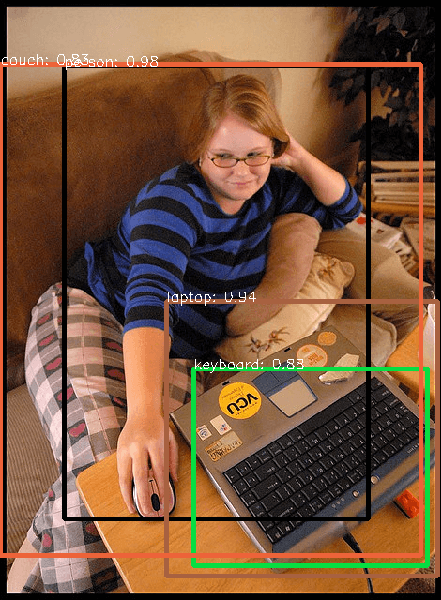}&
		\adjincludegraphics[width=3.1cm,trim={0 {.05\height} 0 {.1\height}},clip,keepaspectratio]{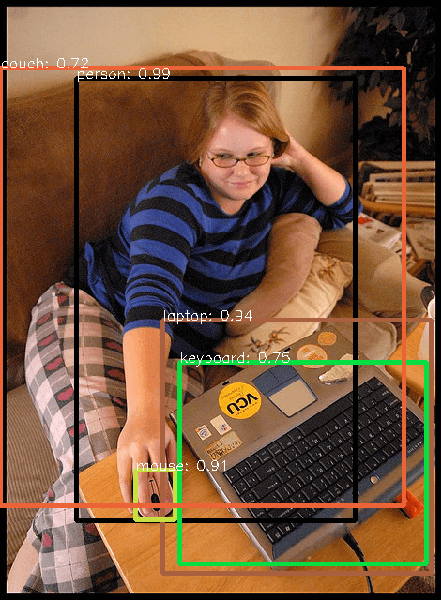}\\
		1NL, BU & 3ENL, TD (ours) & 1NL, BU & 3ENL, TD (ours) \\	
	\end{tabular}
	\vspace{0.2cm}
	\caption{\textbf{Reasoning Module:} Using a relational non-local module directly on the feature maps in a coarse-to-fine manner ease the detection of small objects partly based on the existence of larger related objects. The implicit reasoning process allowing us to locate the drivers in the vehicles and the bat, sandwich, cellphone, clock, mouce toothbrush on the humans hands. Best viewed while zoomed in.}
	\label{fig:reasoning}
\end{figure*}

Figures \ref{fig:small_objects}, \ref{fig:top_down} and \ref{fig:reasoning} show additional examples of interest that emphasize various aspects of our architecture.
We compare between the detection results of our architecture (\textbf{3ENL, TD} with resnet50 FPN backbone, second and forth columns) and the detection results of \cite{wang2017non} (with a non-local module integrated on the higher layers of Faster RCNN (\textbf{1NL, BU}, first and third columns) with the same backbone. Another possible comparison, to Faster RCNN, is possible, but would be less challenging.

Figure \ref{fig:small_objects} shows our improvement in small objects detection due to repeatition in the scene. Our architecture detects more instances and better distiguish between overlapping instances of the same class (see some examples of overlapping instaces in the left most column). The shown examples emphasize the potential in applying non local processing in lower layers of the network, enabled due to the efficient implementation.

Figure \ref{fig:top_down} shows the detection of small objects, partly based on the existence of larger instances of the same class.
Figure \ref{fig:reasoning} demonstrate the implicit reasoning process, carrying information from larger, easy to detect objects to smaller related objects. Typical examples include objects held or carried by humans or drivers in vehicles. These figures emphasize the importance of the top-down modulation in the reasoning process.

\bibliography{egbib}

\begin{thebibliography}{47}
\providecommand{\natexlab}[1]{#1}
\providecommand{\url}[1]{\texttt{#1}}
\expandafter\ifx\csname urlstyle\endcsname\relax
  \providecommand{\doi}[1]{doi: #1}\else
  \providecommand{\doi}{doi: \begingroup \urlstyle{rm}\Url}\fi

\bibitem[Battaglia et~al.(2016)Battaglia, Pascanu, Lai, Rezende,
  et~al.]{battaglia2016interaction}
Peter Battaglia, Razvan Pascanu, Matthew Lai, Danilo~Jimenez Rezende, et~al.
\newblock Interaction networks for learning about objects, relations and
  physics.
\newblock In \emph{Advances in neural information processing systems}, pages
  4502--4510, 2016.

\bibitem[Battaglia et~al.(2018)Battaglia, Hamrick, Bapst, Sanchez-Gonzalez,
  Zambaldi, Malinowski, Tacchetti, Raposo, Santoro, Faulkner,
  et~al.]{battaglia2018relational}
Peter~W Battaglia, Jessica~B Hamrick, Victor Bapst, Alvaro Sanchez-Gonzalez,
  Vinicius Zambaldi, Mateusz Malinowski, Andrea Tacchetti, David Raposo, Adam
  Santoro, Ryan Faulkner, et~al.
\newblock Relational inductive biases, deep learning, and graph networks.
\newblock \emph{arXiv preprint arXiv:1806.01261}, 2018.

\bibitem[Bell et~al.(2016)Bell, Lawrence~Zitnick, Bala, and
  Girshick]{bell2016inside}
Sean Bell, C~Lawrence~Zitnick, Kavita Bala, and Ross Girshick.
\newblock Inside-outside net: Detecting objects in context with skip pooling
  and recurrent neural networks.
\newblock In \emph{Proceedings of the IEEE conference on computer vision and
  pattern recognition}, pages 2874--2883, 2016.

\bibitem[Bronstein et~al.(2017)Bronstein, Bruna, LeCun, Szlam, and
  Vandergheynst]{bronstein2017geometric}
Michael~M Bronstein, Joan Bruna, Yann LeCun, Arthur Szlam, and Pierre
  Vandergheynst.
\newblock Geometric deep learning: going beyond euclidean data.
\newblock \emph{IEEE Signal Processing Magazine}, 34\penalty0 (4):\penalty0
  18--42, 2017.

\bibitem[Cai et~al.(2016)Cai, Fan, Feris, and Vasconcelos]{cai2016unified}
Zhaowei Cai, Quanfu Fan, Rogerio~S Feris, and Nuno Vasconcelos.
\newblock A unified multi-scale deep convolutional neural network for fast
  object detection.
\newblock In \emph{European Conference on Computer Vision}, pages 354--370.
  Springer, 2016.

\bibitem[Chen et~al.(2014)Chen, Papandreou, Kokkinos, Murphy, and
  Yuille]{chen2014semantic}
Liang-Chieh Chen, George Papandreou, Iasonas Kokkinos, Kevin Murphy, and Alan~L
  Yuille.
\newblock Semantic image segmentation with deep convolutional nets and fully
  connected crfs.
\newblock \emph{arXiv preprint arXiv:1412.7062}, 2014.

\bibitem[Chen et~al.(2018)Chen, Li, Fei-Fei, and Gupta]{chen2018iterative}
Xinlei Chen, Li-Jia Li, Li~Fei-Fei, and Abhinav Gupta.
\newblock Iterative visual reasoning beyond convolutions.
\newblock \emph{arXiv preprint arXiv:1803.11189}, 2018.

\bibitem[Deng et~al.(2009)Deng, Dong, Socher, Li, Li, and
  Fei-Fei]{deng2009imagenet}
Jia Deng, Wei Dong, Richard Socher, Li-Jia Li, Kai Li, and Li~Fei-Fei.
\newblock Imagenet: A large-scale hierarchical image database.
\newblock In \emph{Computer Vision and Pattern Recognition, 2009. CVPR 2009.
  IEEE Conference on}, pages 248--255. Ieee, 2009.

\bibitem[Divvala et~al.(2009)Divvala, Hoiem, Hays, Efros, and
  Hebert]{divvala2009empirical}
Santosh~K Divvala, Derek Hoiem, James~H Hays, Alexei~A Efros, and Martial
  Hebert.
\newblock An empirical study of context in object detection.
\newblock In \emph{Computer Vision and Pattern Recognition, 2009. CVPR 2009.
  IEEE Conference on}, pages 1271--1278. IEEE, 2009.

\bibitem[Fu et~al.(2017)Fu, Liu, Ranga, Tyagi, and Berg]{fu2017dssd}
Cheng-Yang Fu, Wei Liu, Ananth Ranga, Ambrish Tyagi, and Alexander~C Berg.
\newblock Dssd: Deconvolutional single shot detector.
\newblock \emph{arXiv preprint arXiv:1701.06659}, 2017.

\bibitem[Galleguillos and Belongie(2010)]{galleguillos2010context}
Carolina Galleguillos and Serge Belongie.
\newblock Context based object categorization: A critical survey.
\newblock \emph{Computer vision and image understanding}, 114\penalty0
  (6):\penalty0 712--722, 2010.

\bibitem[Girshick(2015)]{girshick2015fast}
Ross Girshick.
\newblock Fast r-cnn.
\newblock In \emph{Proceedings of the IEEE international conference on computer
  vision}, pages 1440--1448, 2015.

\bibitem[Girshick et~al.(2014)Girshick, Donahue, Darrell, and
  Malik]{girshick2014rich}
Ross Girshick, Jeff Donahue, Trevor Darrell, and Jitendra Malik.
\newblock Rich feature hierarchies for accurate object detection and semantic
  segmentation.
\newblock In \emph{Proceedings of the IEEE conference on computer vision and
  pattern recognition}, pages 580--587, 2014.

\bibitem[Gori et~al.(2005)Gori, Monfardini, and Scarselli]{gori2005new}
Marco Gori, Gabriele Monfardini, and Franco Scarselli.
\newblock A new model for learning in graph domains.
\newblock In \emph{Neural Networks, 2005. IJCNN'05. Proceedings. 2005 IEEE
  International Joint Conference on}, volume~2, pages 729--734. IEEE, 2005.

\bibitem[Goyal et~al.(2017)Goyal, Doll{\'a}r, Girshick, Noordhuis, Wesolowski,
  Kyrola, Tulloch, Jia, and He]{goyal2017accurate}
Priya Goyal, Piotr Doll{\'a}r, Ross Girshick, Pieter Noordhuis, Lukasz
  Wesolowski, Aapo Kyrola, Andrew Tulloch, Yangqing Jia, and Kaiming He.
\newblock Accurate, large minibatch sgd: training imagenet in 1 hour.
\newblock \emph{arXiv preprint arXiv:1706.02677}, 2017.

\bibitem[Hariharan et~al.(2017)Hariharan, Arbelaez, Girshick, and
  Malik]{hariharan2017object}
Bharath Hariharan, Pablo Arbelaez, Ross Girshick, and Jitendra Malik.
\newblock Object instance segmentation and fine-grained localization using
  hypercolumns.
\newblock \emph{IEEE Transactions on Pattern Analysis \& Machine Intelligence},
  4:\penalty0 627--639, 2017.

\bibitem[He et~al.(2014)He, Zhang, Ren, and Sun]{he2014spatial}
Kaiming He, Xiangyu Zhang, Shaoqing Ren, and Jian Sun.
\newblock Spatial pyramid pooling in deep convolutional networks for visual
  recognition.
\newblock In \emph{European conference on computer vision}, pages 346--361.
  Springer, 2014.

\bibitem[He et~al.(2017)He, Gkioxari, Doll{\'a}r, and Girshick]{he2017mask}
Kaiming He, Georgia Gkioxari, Piotr Doll{\'a}r, and Ross Girshick.
\newblock Mask r-cnn.
\newblock In \emph{Computer Vision (ICCV), 2017 IEEE International Conference
  on}, pages 2980--2988. IEEE, 2017.

\bibitem[Hu et~al.(2017)Hu, Gu, Zhang, Dai, and Wei]{hu2017relation}
Han Hu, Jiayuan Gu, Zheng Zhang, Jifeng Dai, and Yichen Wei.
\newblock Relation networks for object detection.
\newblock \emph{arXiv preprint arXiv:1711.11575}, 8, 2017.

\bibitem[Johnson et~al.(2015)Johnson, Krishna, Stark, Li, Shamma, Bernstein,
  and Fei-Fei]{johnson2015image}
Justin Johnson, Ranjay Krishna, Michael Stark, Li-Jia Li, David Shamma, Michael
  Bernstein, and Li~Fei-Fei.
\newblock Image retrieval using scene graphs.
\newblock In \emph{Proceedings of the IEEE conference on computer vision and
  pattern recognition}, pages 3668--3678, 2015.

\bibitem[Kong et~al.(2016)Kong, Yao, Chen, and Sun]{kong2016hypernet}
Tao Kong, Anbang Yao, Yurong Chen, and Fuchun Sun.
\newblock Hypernet: Towards accurate region proposal generation and joint
  object detection.
\newblock In \emph{Proceedings of the IEEE conference on computer vision and
  pattern recognition}, pages 845--853, 2016.

\bibitem[Kr{\"a}henb{\"u}hl and Koltun(2011)]{krahenbuhl2011efficient}
Philipp Kr{\"a}henb{\"u}hl and Vladlen Koltun.
\newblock Efficient inference in fully connected crfs with gaussian edge
  potentials.
\newblock In \emph{Advances in neural information processing systems}, pages
  109--117, 2011.

\bibitem[Lafferty et~al.(2001)Lafferty, McCallum, and
  Pereira]{lafferty2001conditional}
John Lafferty, Andrew McCallum, and Fernando~CN Pereira.
\newblock Conditional random fields: Probabilistic models for segmenting and
  labeling sequence data.
\newblock 2001.

\bibitem[Lin et~al.(2014{\natexlab{a}})Lin, Maire, Belongie, Hays, Perona,
  Ramanan, Dollar, and Zitnick]{coco_det_eval}
Tsung-Yi Lin, Michael Maire, Serge Belongie, James Hays, Pietro Perona, Deva
  Ramanan, Piotr Dollar, and C~Lawrence Zitnick.
\newblock Microsoft coco: Detection evaluation, 2014{\natexlab{a}}.
\newblock URL \url{http://cocodataset.org/detection-eval}.

\bibitem[Lin et~al.(2014{\natexlab{b}})Lin, Maire, Belongie, Hays, Perona,
  Ramanan, Doll{\'a}r, and Zitnick]{lin2014microsoft}
Tsung-Yi Lin, Michael Maire, Serge Belongie, James Hays, Pietro Perona, Deva
  Ramanan, Piotr Doll{\'a}r, and C~Lawrence Zitnick.
\newblock Microsoft coco: Common objects in context.
\newblock In \emph{European conference on computer vision}, pages 740--755.
  Springer, 2014{\natexlab{b}}.

\bibitem[Lin et~al.(2017)Lin, Doll{\'a}r, Girshick, He, Hariharan, and
  Belongie]{lin2017feature}
Tsung-Yi Lin, Piotr Doll{\'a}r, Ross~B Girshick, Kaiming He, Bharath Hariharan,
  and Serge~J Belongie.
\newblock Feature pyramid networks for object detection.
\newblock In \emph{CVPR}, volume~1, page~4, 2017.

\bibitem[Liu et~al.(2018)Liu, Wen, Fan, Loy, and Huang]{liu2018non}
Ding Liu, Bihan Wen, Yuchen Fan, Chen~Change Loy, and Thomas~S Huang.
\newblock Non-local recurrent network for image restoration.
\newblock \emph{arXiv preprint arXiv:1806.02919}, 2018.

\bibitem[Liu et~al.(2017)Liu, Huang, and Wang]{liu2017receptive}
Songtao Liu, Di~Huang, and Yunhong Wang.
\newblock Receptive field block net for accurate and fast object detection.
\newblock \emph{arXiv preprint arXiv:1711.07767}, 2017.

\bibitem[Liu et~al.(2016)Liu, Anguelov, Erhan, Szegedy, Reed, Fu, and
  Berg]{liu2016ssd}
Wei Liu, Dragomir Anguelov, Dumitru Erhan, Christian Szegedy, Scott Reed,
  Cheng-Yang Fu, and Alexander~C Berg.
\newblock Ssd: Single shot multibox detector.
\newblock In \emph{European conference on computer vision}, pages 21--37.
  Springer, 2016.

\bibitem[Martucci(1994)]{martucci1994symmetric}
Stephen~A Martucci.
\newblock Symmetric convolution and the discrete sine and cosine transforms.
\newblock \emph{IEEE Transactions on Signal Processing}, 42\penalty0
  (5):\penalty0 1038--1051, 1994.

\bibitem[Massa and Girshick(2018)]{massa2018mrcnn}
Francisco Massa and Ross Girshick.
\newblock maskrcnn-benchmark: Fast, modular reference implementation of
  instance segmentation and object detection algorithms in pytorch.
\newblock \url{https://github.com/facebookresearch/maskrcnn-benchmark}, 2018.

\bibitem[Newell et~al.(2016)Newell, Yang, and Deng]{newell2016stacked}
Alejandro Newell, Kaiyu Yang, and Jia Deng.
\newblock Stacked hourglass networks for human pose estimation.
\newblock In \emph{European Conference on Computer Vision}, pages 483--499.
  Springer, 2016.

\bibitem[Raposo et~al.(2017)Raposo, Santoro, Barrett, Pascanu, Lillicrap, and
  Battaglia]{raposo2017discovering}
David Raposo, Adam Santoro, David Barrett, Razvan Pascanu, Timothy Lillicrap,
  and Peter Battaglia.
\newblock Discovering objects and their relations from entangled scene
  representations.
\newblock \emph{arXiv preprint arXiv:1702.05068}, 2017.

\bibitem[Redmon and Farhadi(2018)]{redmon2018yolov3}
Joseph Redmon and Ali Farhadi.
\newblock Yolov3: An incremental improvement.
\newblock \emph{arXiv preprint arXiv:1804.02767}, 2018.

\bibitem[Redmon et~al.(2016)Redmon, Divvala, Girshick, and
  Farhadi]{redmon2016you}
Joseph Redmon, Santosh Divvala, Ross Girshick, and Ali Farhadi.
\newblock You only look once: Unified, real-time object detection.
\newblock In \emph{Proceedings of the IEEE conference on computer vision and
  pattern recognition}, pages 779--788, 2016.

\bibitem[Ren et~al.(2017)Ren, He, Girshick, and Sun]{ren2017faster}
Shaoqing Ren, Kaiming He, Ross Girshick, and Jian Sun.
\newblock Faster r-cnn: towards real-time object detection with region proposal
  networks.
\newblock \emph{IEEE Transactions on Pattern Analysis \& Machine Intelligence},
  6:\penalty0 1137--1149, 2017.

\bibitem[Ronneberger et~al.(2015)Ronneberger, Fischer, and
  Brox]{ronneberger2015u}
Olaf Ronneberger, Philipp Fischer, and Thomas Brox.
\newblock U-net: Convolutional networks for biomedical image segmentation.
\newblock In \emph{International Conference on Medical image computing and
  computer-assisted intervention}, pages 234--241. Springer, 2015.

\bibitem[Santoro et~al.(2017)Santoro, Raposo, Barrett, Malinowski, Pascanu,
  Battaglia, and Lillicrap]{santoro2017simple}
Adam Santoro, David Raposo, David~G Barrett, Mateusz Malinowski, Razvan
  Pascanu, Peter Battaglia, and Tim Lillicrap.
\newblock A simple neural network module for relational reasoning.
\newblock In \emph{Advances in neural information processing systems}, pages
  4967--4976, 2017.

\bibitem[Scarselli et~al.(2009)Scarselli, Gori, Tsoi, Hagenbuchner, and
  Monfardini]{scarselli2009graph}
Franco Scarselli, Marco Gori, Ah~Chung Tsoi, Markus Hagenbuchner, and Gabriele
  Monfardini.
\newblock The graph neural network model.
\newblock \emph{IEEE Transactions on Neural Networks}, 20\penalty0
  (1):\penalty0 61--80, 2009.

\bibitem[Shen et~al.(2017)Shen, Liu, Li, Jiang, Chen, and Xue]{shen2017dsod}
Zhiqiang Shen, Zhuang Liu, Jianguo Li, Yu-Gang Jiang, Yurong Chen, and
  Xiangyang Xue.
\newblock Dsod: Learning deeply supervised object detectors from scratch.
\newblock In \emph{The IEEE International Conference on Computer Vision
  (ICCV)}, volume~3, page~7, 2017.

\bibitem[Shrivastava et~al.(2016)Shrivastava, Sukthankar, Malik, and
  Gupta]{shrivastava2016beyond}
Abhinav Shrivastava, Rahul Sukthankar, Jitendra Malik, and Abhinav Gupta.
\newblock Beyond skip connections: Top-down modulation for object detection.
\newblock \emph{arXiv preprint arXiv:1612.06851}, 2016.

\bibitem[Tomasi and Manduchi(1998)]{tomasi1998bilateral}
Carlo Tomasi and Roberto Manduchi.
\newblock Bilateral filtering for gray and color images.
\newblock In \emph{Iccv}, volume~98, page~2, 1998.

\bibitem[Vaswani et~al.(2017)Vaswani, Shazeer, Parmar, Uszkoreit, Jones, Gomez,
  Kaiser, and Polosukhin]{vaswani2017attention}
Ashish Vaswani, Noam Shazeer, Niki Parmar, Jakob Uszkoreit, Llion Jones,
  Aidan~N Gomez, {\L}ukasz Kaiser, and Illia Polosukhin.
\newblock Attention is all you need.
\newblock In \emph{Advances in Neural Information Processing Systems}, pages
  5998--6008, 2017.

\bibitem[Wang et~al.(2017)Wang, Girshick, Gupta, and He]{wang2017non}
Xiaolong Wang, Ross Girshick, Abhinav Gupta, and Kaiming He.
\newblock Non-local neural networks.
\newblock \emph{arXiv preprint arXiv:1711.07971}, 10, 2017.

\bibitem[Yuan and Wang(2018)]{yuan2018ocnet}
Yuhui Yuan and Jingdong Wang.
\newblock Ocnet: Object context network for scene parsing.
\newblock \emph{arXiv preprint arXiv:1809.00916}, 2018.

\bibitem[Yue et~al.(2018)Yue, Sun, Yuan, Zhou, Ding, and Xu]{yue2018compact}
Kaiyu Yue, Ming Sun, Yuchen Yuan, Feng Zhou, Errui Ding, and Fuxin Xu.
\newblock Compact generalized non-local network.
\newblock In \emph{Advances in Neural Information Processing Systems}, pages
  6510--6519, 2018.

\bibitem[Zhang et~al.(2018)Zhang, Goodfellow, Metaxas, and
  Odena]{zhang2018self}
Han Zhang, Ian Goodfellow, Dimitris Metaxas, and Augustus Odena.
\newblock Self-attention generative adversarial networks.
\newblock \emph{arXiv preprint arXiv:1805.08318}, 2018.

\end{thebibliography}
\end{document}